\documentclass[3p,,preprint,12pt]{elsarticle}
\makeatletter\if@twocolumn\PassOptionsToPackage{switch}{lineno}\else\fi\makeatother

\usepackage{tabulary,xcolor}
\usepackage{amsfonts,amsmath,amssymb}
\usepackage[T1]{fontenc}
\makeatletter
\let\save@ps@pprintTitle\ps@pprintTitle
\def\ps@pprintTitle{\save@ps@pprintTitle\gdef\@oddfoot{\footnotesize\itshape \null\hfill}}
\def\hlinewd#1{%
	\noalign{\ifnum0=`}\fi\hrule \@height #1%
	\futurelet\reserved@a\@xhline}

\def\tblbottomrule{\hlinewd{.8pt}}
\def\tblmidrule{\noalign{\vspace*{6pt}}\hline\noalign{\vspace*{2pt}}}
\AtBeginDocument{\ifNAT@numbers \biboptions{sort&compress}\fi}
\makeatother

\usepackage{ifluatex}
\ifluatex
\usepackage{fontspec}
\defaultfontfeatures{Ligatures=TeX}
\usepackage[]{unicode-math}
\unimathsetup{math-style=TeX}
\else 
\usepackage[utf8]{inputenc}
\fi 
\ifluatex\else\usepackage{stmaryrd}\fi

\usepackage{url,multirow,morefloats,floatflt,cancel,tfrupee}
\makeatletter

\AtBeginDocument{\@ifpackageloaded{textcomp}{}{\usepackage{textcomp}}}
\makeatother
\usepackage{colortbl}
\usepackage{xcolor}
\usepackage{pifont}
\usepackage[nointegrals]{wasysym}
\makeatletter

\def\mcWidth#1{\csname TY@F#1\endcsname+\tabcolsep}

\def\cAlignHack{\rightskip\@flushglue\leftskip\@flushglue\parindent\z@\parfillskip\z@skip}
\def\rAlignHack{\rightskip\z@skip\leftskip\@flushglue \parindent\z@\parfillskip\z@skip}

\if@twocolumn\usepackage{dblfloatfix}\fi 
\AtBeginDocument{
	\expandafter\ifx\csname eqalign\endcsname\relax
	\def\eqalign#1{\null\vcenter{\def\\{\cr}\openup\jot\m@th
			\ialign{\strut$\displaystyle{##}$\hfil&$\displaystyle{{}##}$\hfil
				\crcr#1\crcr}}\,}
	\fi
}

\AtBeginDocument{%
	\@ifpackageloaded{endfloat}%
	{\renewcommand\efloat@iwrite[1]{\immediate\expandafter\protected@write\csname efloat@post#1\endcsname{}}}{}%
}%

\let\lt=<
\let\gt=>
\def\processVert{\ifmmode|\else\textbar\fi}

\@ifundefined{subparagraph}{
	\def\subparagraph{\@startsection{paragraph}{5}{2\parindent}{0ex plus 0.1ex minus 0.1ex}%
		{0ex}{\normalfont\small\itshape}}%
}{}

\newcommand\role[1]{\unskip}
\newcommand\aucollab[1]{\unskip}

\@ifundefined{tsGraphicsScaleX}{\gdef\tsGraphicsScaleX{1}}{}
\@ifundefined{tsGraphicsScaleY}{\gdef\tsGraphicsScaleY{.9}}{}
\def\checkGraphicsWidth{\ifdim\Gin@nat@width>\linewidth
	\tsGraphicsScaleX\linewidth\else\Gin@nat@width\fi}

\def\checkGraphicsHeight{\ifdim\Gin@nat@height>.9\textheight
	\tsGraphicsScaleY\textheight\else\Gin@nat@height\fi}

\def\fixFloatSize#1{}%\@ifundefined{processdelayedfloats}{\setbox0=\hbox{\includegraphics{#1}}\ifnum\wd0<\columnwidth\relax\renewenvironment{figure*}{\begin{figure}}{\end{figure}}\fi}{}}
\let\ts@includegraphics\includegraphics

\def\inlinegraphic[#1]#2{{\edef\@tempa{#1}\edef\baseline@shift{\ifx\@tempa\@empty0\else#1\fi}\edef\tempZ{\the\numexpr(\numexpr(\baseline@shift*\f@size/100))}\protect\raisebox{\tempZ pt}{\ts@includegraphics{#2}}}}

\AtBeginDocument{\def\includegraphics{\@ifnextchar[{\ts@includegraphics}{\ts@includegraphics[width=\checkGraphicsWidth,height=\checkGraphicsHeight,keepaspectratio]}}}

\def\URL#1#2{\@ifundefined{href}{#2}{\href{#1}{#2}}}

\def\UrlOrds{\do\*\do\-\do\~\do\'\do\"\do\-}%
\g@addto@macro{\UrlBreaks}{\UrlOrds}
\@ifundefined{quoteAttrib}
{}
{}
\makeatother

\usepackage{float}
\usepackage{graphicx}

\begin{document}

\begin{frontmatter}
	
\title{Enhanced adaptive cross-layer scheme for low latency HEVC streaming over Vehicular Ad-hoc Networks (VANETs)}
    
\author[afff8eb478d98a82f299fd2d7a135377b8c,afff34f8e25ad1947f358f332e7e95cdc3f]{Mohamed Aymen Labiod \corref{cor1}}
\ead{Mohamed-Aymen.Labiod@u-pec.fr}
\cortext[cor1]{Corresponding author}
\author[afff8eb478d98a82f299fd2d7a135377b8c]{Mohamed Gharbi}
\ead{Mohamed.Gharbi@uphf.fr}
\author[afff8eb478d98a82f299fd2d7a135377b8c]{Fran\c{c}ois-Xavier Coudoux}
\ead{Francois-Xavier.Coudoux@uphf.fr}
\author[afff8eb478d98a82f299fd2d7a135377b8c]{Patrick Corlay}
\ead{Patrick.Corlay@uphf.fr}
\author[afff34f8e25ad1947f358f332e7e95cdc3f]{Noureddine Doghmane}
\ead{ndoghmane@univ-annaba.org}

\address[afff8eb478d98a82f299fd2d7a135377b8c]{
    Univ. Polytechnique Hauts-de-France, CNRS, Univ. Lille, YNCREA, Centrale Lille, UMR 8520 - IEMN, DOAE, F-59313 Valenciennes, France}
  	
\address[afff34f8e25ad1947f358f332e7e95cdc3f]{
    Automatic and signals laboratory Annaba (LASA), Department of electronic, Badji Mokhtar University Annaba, Algeria}

\begin{abstract}
Vehicular communication has become a reality guided by various applications. Among those, high video quality delivery with low latency constraints required by real-time applications constitutes a very challenging task. By dint of its never-before-achieved compression level, the new High-Efficiency Video Coding (HEVC) is very promising for real-time video streaming through Vehicular Ad-hoc Networks (VANET). However, these networks have variable channel quality and limited bandwidth. Therefore, ensuring satisfactory video quality on such networks is a major challenge. In this work, a low complexity cross-layer mechanism is proposed to improve end-to-end performances of HEVC video streaming in VANET under low delay constraints. The idea is to assign to each packet of the transmitted video the most appropriate Access Category (AC) queue on the Medium Access Control (MAC) layer, considering the temporal prediction structure of the video encoding process, the importance of the frame and the state of the network traffic load. Simulation results demonstrate that for different targeted low-delay video communication scenarios, the proposed mechanism offers significant improvements regarding video quality at the reception and end-to-end delay compared to the Enhanced Distributed Channel Access (EDCA) adopted in the 802.11p. Both Quality of Service (QoS) and Quality of Experience (QoE) evaluations have been also carried out to validate the proposed approach.
\end{abstract}
\end{frontmatter}
    
\section{Introduction}
In recent years, vehicular communication and intelligent transportation systems (ITS) have attracted the interest of the scientific and professional communities. The development of VANETs has raised the possibility of deploying applications aimed at security, traffic management as well as comfort and entertainments. The transmission of video contents over vehicular networks would represent a further advancement \unskip~\cite{junior_game_2018}. Indeed, visual information data is needed for many applications like overtaking maneuver, pedestrian crossing assistance, public transport assistance, video surveillance and video communication for entertainments \unskip~\cite{jiau_multimedia_2015, gerla_content_2014}. However, the compressed video is highly sensitive to channel noise and losses. Moreover, VANETs suffer from severe transmission conditions and an associated packet loss rate (PLR) that provides no QoS guarantees.

Several technological solutions have been proposed to enhance multimedia transmissions over vehicular networks \unskip~\cite{campolo_todays_2015}. In particular, the IEEE 802.11p standard dedicated to vehicular networks copes with the varying QoS by providing differentiated classes of service at the MAC layer \unskip~\cite{259927:5821247}. On the other hand, the HEVC/H265, has recently been developed, which outperforms its predecessor (H264/AVC) in coding efficiency of about 50\% \unskip~\cite{259927:5821246}. In vehicular applications requiring video, such as monitoring and road traffic optimization, the guarantee of low delay is essential \unskip~\cite{parvez_survey_2018, quadros_qoe-driven_2016}. It is even more vital in driver-assistance systems \unskip~\cite{6170893} and vehicles remote control applications, considering the recent interest in autonomous vehicles. Therefore, communication systems must guarantee low latency while guaranteeing high reliability \unskip~\cite{alieiev_automotive_2015}. In order to improve the multimedia transmission and to ensure an acceptable video quality service, it is possible to optimize the parameters of the QoS. 

In a vehicular environment, received signal power may undergo wide variations which are due to several factors, mainly fading, shadowing, multipath propagation and Doppler effect. Hence, VANETs are judged as a network with harsh channel conditions that leads to a degradation of the link throughput and consequently poor video quality. For this purpose, many studies have done an evaluation of the video quality relating to either the network load \unskip~\cite{259927:5821245} or according to the video source encoder \unskip~\cite{259927:5821244}. Torres et al. \unskip~\cite{Torres} proposed a performance evaluation for real-time video transmission in vehicular environments. The study focused on the influence of vehicle distance and vehicular density on HEVC video encoded sequence for highway and urban scenarios. The packet delivery ratio (PDR) and the peak signal to noise ratio (PSNR) were used as evaluation metrics. In order to improve this transmission, several solutions have also been proposed on various aspects of the transmission chain. Mammeri et al. \unskip~\cite{259927:5821242} proposed a modification of the real-time transport protocol (RTP) to facilitate the transmission of video encoded in H.264. Video transmission implementation in VANET has also been realized. While Zaidi et al. \unskip~\cite{259927:5821241} have proposed an error recovery protocol, based on a retransmission technique. The protocol uses video frames unequal protection for MPEG4 part 2 video encoding. As for Rezende et al. \unskip~\cite{259927:5821243} they used network coding and erasure coding to video streaming over VANETs.

On the other hand, regardless of the video transmission, some improvements have been made in the MAC protocols for safety applications in VANET. Gupta et al. \unskip~\cite{259927:5821240} presented a comprehensive state-of-the-art survey of some of these works. While the survey \unskip~\cite{259927:5821239} gives a detailed overview of crosslayer design strategies and the challenges associated with them in VANETs. The survey explores different cross-layer schemes across the different open systems interconnection model (OSI model) layers.

Improvements were also made on the EDCA in the case of video transmission over IEEE 802.11e standard. Ksentini et al. \unskip~\cite{259927:5821238} were the forerunners with the idea of using the others access categories (AC) dedicated to lower priority flows, i.e. best effort (BE) and background traffic (BK), that the EDCA makes available according to the video coding significance. The authors proposed a cross-layer architecture to improve H.264 video transmission over an IEEE 802.11e network using a mapping algorithm, based on the traffic specification of IEEE 802.11e EDCA. However, this mapping is static and does not consider the state of the network. Lin et al. \unskip~\cite{259927:5821237} proposed an adaptive cross-layer mapping algorithm to improve the video delivery quality of MPEG-4 Part 2 video over IEEE 802.11e wireless networks. The improvement in the quality of the delivered video by exploiting the hierarchical characteristics of video frames and passing video data significance information from the application layer to the MAC layer in a cross-layer design architecture. Moreover, the mapping of the video packet to the appropriate AC is based on the state of the network traffic and the importance of the video frame. Ke et al. \unskip~\cite{259927:5821236} have proposed another algorithm in which they added a consideration of the actions performed on the previous packets of the same video frame.

Mai et al. \unskip~\cite{259927:5821235} implement a cross-layer system for H.264/AVC video streaming over IEEE 802.11e wireless networks. The proposed mechanism makes more efficient use of the radio source in estimating the access waiting time of each AC by selecting for each packet the AC destination with the smallest expected access waiting time. However, works listed for cross-layer methods are specific to IEEE 802.11e and based on older video encoding standards. These works do not consider the change of temporal prediction structure that the video encoder may induce.

Also, they do not take into account the latency issue for a low-delay transmission. Indeed, the video encoders used, or the techniques proposed require a fairly significant latency dedicated for delay insensitive applications. Chen et al. \unskip~\cite{259927:5821234} have established a delay-rate-distortion model in a wireless video communication system using the Low Delay (LD) mode of H.264. In this mode, the group of pictures (GoP) structure is composed of intra and predicted frames known as I- and P-frames respectively; since in real-time applications, the delay constraint does not allow the use of bidirectional predicted frames, also known as B-frames. The proposed model makes it possible to establish a functional relationship between the video encoding time, the rate in the video encoder and the distortion of the source. While Kokkonis et al. \unskip~\cite{259927:5821233} proposed an algorithm for the real-time transmission of a H.265/HEVC stream with haptic data over the Internet. The algorithm chooses the optimal temporal prediction to be used taking into account the HEVC encoding and decoding times and also the network QoS. They further showed the influence of frame coding order within a GoP on end-to-end delay and undertook comparative tests between H.264 and HEVC. Vinel et al. \unskip~\cite{259927:5821232} provides a real-time codec channel adaptation approach in a video transmission over IEEE 802.11p standard. The end-to-end model is proposed for video-based overtaking assistance application with maximum end-to-end latency that should not exceed 200 ms as estimated.

In the present work, a cross-layer mechanism is proposed to improve HEVC video streaming in VANETs under low delay constraint. We propose two mapping mechanisms dedicated to the IEEE 802.11p standard aiming to see the efficiency of these methods in the context of vehicular networks with varying network topology. The first mapping algorithm is static, while the second is adaptive according to the video temporal prediction structure. Moreover, the algorithms take advantage of the new temporal prediction structures introduced in HEVC. Knowing the importance of low latency transmission in vehicular networks. The mechanisms initially proposed in \unskip~\cite{259927:5821238} and \unskip~\cite{259927:5821237} have been redesigned in the context of both HEVC and IEEE 802.11p standards. The proposed mechanisms consider both the importance of the frame in a video and the state of the channel being determined by the queue length of the MAC layer. The idea is to assign each packet of the transmitted video to the most appropriate AC queue on the MAC layer, considering the temporal prediction structure of the video, the importance of each frame and the state of the instantaneous network traffic load. The proposed cross-layer mechanism was evaluated by simulation in both highway and suburban realistic vehicular environments.  

The remainder of this paper is organized as follows. Section 2 presents a description of the proposed solution. Section 3, concerns the framework and the simulations conducted. Simulation results are presented in Section 4. They demonstrate the proposed solutions effectiveness, providing 17\% average received packets gain compared to the IEEE 802.11p EDCA mechanism. Finally, a conclusion is discussed in Section 5.

\section{Description of the proposed solution}
Cross-layer design refers to a mechanism that exploits the dependency between protocol layers to achieve significant performance gains. Several design types can be listed, depending on how information is exchanged between layers. Srivastava et al. \unskip~\cite{259927:5821231} reduced the possible designs to four different approaches. The first approach consists in the creation of new interfaces, the second in the merging of adjacent layers, the third one in the design coupling without new interfaces and finally the last one in a vertical calibration across the layers.  

The proposed cross-layer architecture uses information about the significance of video packets from the application layer to exploit it on decision process at the MAC layer. In the rest of this section, we will describe the technologies used in this work. In the first place, we will explain the characteristics of the IEEE 802.11p standard specific to vehicular networks. In the second place, we will provide a brief overview of H.265/HEVC encoding, and we will conclude this section with presenting our proposed cross-layer system.

\subsection{The IEEE 802.11p standard }The IEEE 802.11p standard is an approved amendment to the IEEE 802.11 standard to provide wireless access in vehicular environment (WAVE). The standard uses a modified version of IEEE 802.11a for the PHY layer based on dedicated short-range communication (DSRC) standard in the 5.850-5.925 GHz frequency range. DSRC is considered to provide communication for both vehicular to vehicular (V2V) and for vehicular to infrastructure (V2I) scenarios \unskip~\cite{259927:5821230}. The IEEE 802.p being dedicated to the United States, the European Standard Telecommunications Institute (ETSI) defines ITS-G5 as its equivalent standard in Europe \unskip~\cite{noauthor_intelligent_2013}. Some differences between the two standards exist at the upper layers. Nevertheless, like the DSRC, it also works in the 5.9 GHz band \unskip~\cite{festag_standards_2015}. In Japan, the DSRC equivalent is used in the 5.8 GHz spectrum.

The DSRC frequency spectrum is composed of six service channels (SCH) and one control channel (CCH). It allows data rates of 3, 6, 9, 12, 18, 24 and 27 Mbps with a preamble of 3 Mbps. The modulation used is the Orthogonal Frequency Division Multiplexing (OFDM).

The medium access control layer protocol of the IEEE 802.11p standard uses CSMA/CA (Carrier Sense Multiple Access with Collision Avoidance) as basic medium access scheme for link sharing, and EDCA for packet transmission \unskip~\cite{259927:5821229}. The EDCA protocol provides service hierarchization using flow prioritization according to QoS requirements \unskip~\cite{259927:5821228}. It has been initially introduced in the IEEE.802.11e standard and has undergone some modifications \unskip~\cite{259927:5821240,259927:5821227}.

Indeed, EDCA is an improvement of the distributed channel access (DCA) method to provide required QoS. Instead of single queue storing data frame, EDCA uses four queues representing different levels of priority or access categories noted hereafter ACs. Each of these ACs is dedicated to a kind of traffic, namely background (BK or AC0), best effort (BE or AC1), video (VI or AC2) and voice (VO or AC3) as illustrated in Figure~\ref{figure-08dc6f107ddbbdf435f0d72bf237cbc5}. 

\bgroup
\fixFloatSize{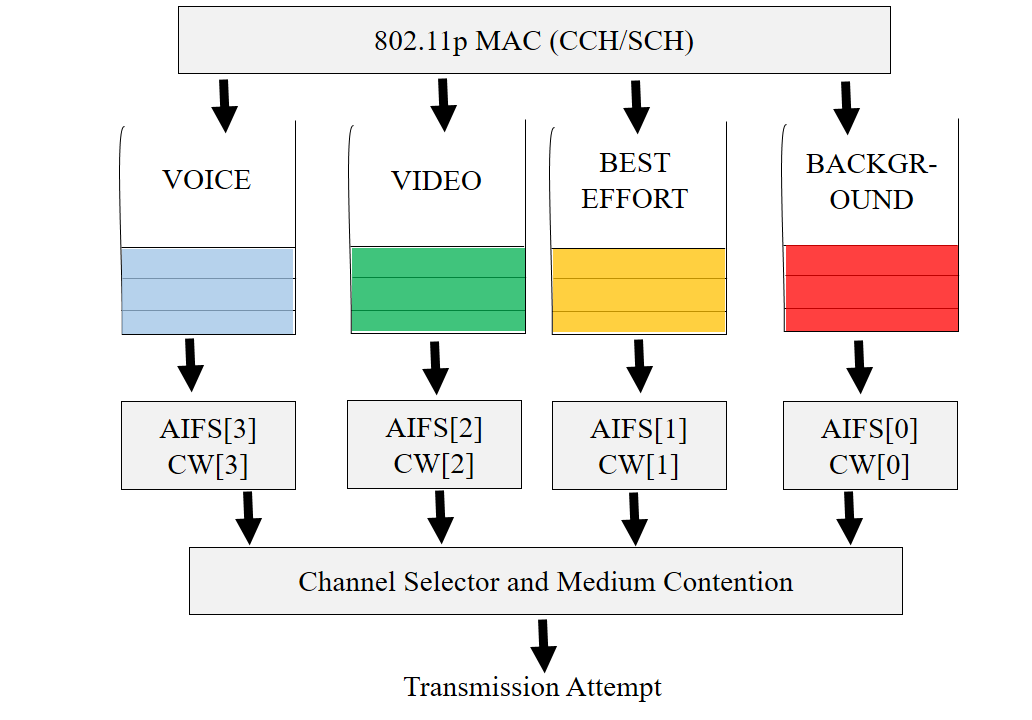}
\begin{figure*}[!htbp]
\centering \makeatletter\IfFileExists{images/Figure1.png}{\includegraphics[width=.78\linewidth]{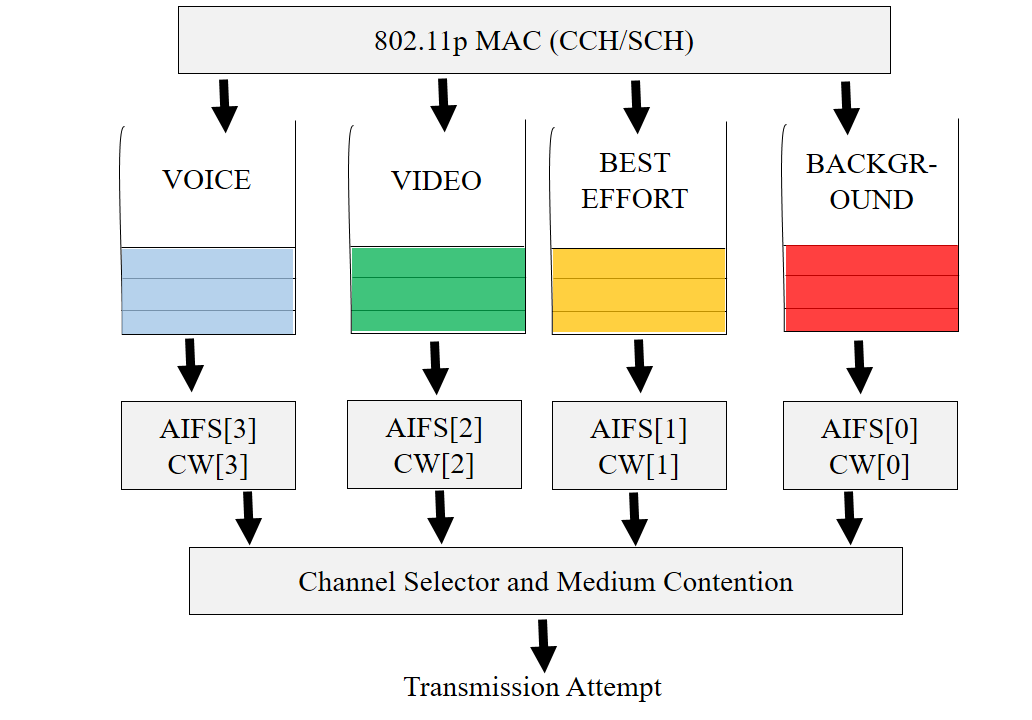}}{}
\makeatother 
\caption{{The different access categories in the IEEE 802.11p MAC architecture.}}
\label{figure-08dc6f107ddbbdf435f0d72bf237cbc5}
\end{figure*}
\egroup
Higher is the transmission priority, better is the transmitting possibility. For this purpose, priority is assigned according to the importance of each traffic. Highest priority was given to voice AC then comes video traffic, best effort, and background traffic respectively. 

The prioritization is established according to the waiting time \textit{T\ensuremath{_{\rm AIFS\ }}} (Time Arbitration Inter-Frame Space), representing the time needed to access the medium for each AC. It allows a different prioritization of the frames depending on the traffic type. For example, a short \textit{T\ensuremath{_{\rm AIFS}}} will shorten the time between two frames and especially a shorter access time to the medium. The \textit{T\ensuremath{_{\rm AIFS}}} value is given by \unskip~\cite{259927:5821240}:

\begin{equation}
{T_{AIFS}}[AC]=AIFSN[AC]*aSlotTime+SIFS
\label{moneq}
\end{equation}

where the AC is the AC of each traffic type, \textit{AIFSN [AC]} (Arbitration Inter-Frame Space Number) is the predefined constant corresponding to each AC. The Short Inter-Frame Space (\textit{SIFS}) and \textit{aSlotTime} are constant intervals predefined in the standard defined respectively at 32 \ensuremath{\mu }s and 13 \ensuremath{\mu }s. One other difference between the ACs is the contention windows (CW). 

Since EDCA has four queues, internal collisions between queues can occur. The mechanism cited before helps solve these problems. Figure~\ref{figure-b0dda45e191a46e975d0b852637b85d7} illustrates an example of contending for access to the medium and the prioritization established by \textit{T\ensuremath{_{\rm AIFS}}}. We can see that a voice frame and a best effort frame are in competition to have access to the medium. The lower waiting time of the AC voice allows it to access to the medium at the expense of the best effort. The values of each AC are given in Table~\ref{table-wrap-e4d6724dea32374aa7fa31c0b61a4a2b}\unskip~\cite{259927:5821240}. Different AIFSN and CW values are chosen for the different types of ACs in the CCH and SCH. We verify that video AC priority is higher than those of the BE and BK represented by a smaller \textit{T\ensuremath{_{\rm AIFS}}}.

\begin{table*}[!htbp]
	\caption{{ IEEE 802.11p Access Categories. } }
	\label{table-wrap-e4d6724dea32374aa7fa31c0b61a4a2b}
	\def\arraystretch{1}
	\ignorespaces 
	\centering 
	\resizebox{\textwidth}{!}{%
		\begin{tabulary}{\linewidth}{llllllllll}
			\hline 
			\multirow{2}{*}{\begin{tabular}[c]{@{}c@{}}AC \\ Number\end{tabular}} & \multirow{2}{*}{\begin{tabular}[c]{@{}c@{}}Access\\ Category\end{tabular}} & \multicolumn{4}{c}{CCH}            & \multicolumn{4}{c}{SCH}            \\
			&                                                                            & CWmin & CWmax & AIFSN & {\begin{tabular}[c]{@{}c@{}}{T\ensuremath{_{\rm AIFS}}} \\ (µs)\end{tabular}}& CWmin & CWmax & AIFSN & {\begin{tabular}[c]{@{}c@{}}{T\ensuremath{_{\rm AIFS}}}  \\ (µs)\end{tabular}} \\
			\tblmidrule 
			0                                                                     & \begin{tabular}[c]{@{}c@{}}Background\\ Traffic (BK)\end{tabular}          & 15    & 1023  & 9     & 149        & 31    & 1023  & 7     & 123        \\
			1                                                                     & \begin{tabular}[c]{@{}c@{}}Best Effort \\ (BE)\end{tabular}                & 7     & 15    & 6     & 110        & 31    & 1023  & 3     & 71         \\
			2                                                                     & Video(VI)                                                                  & 3     & 7     & 3     & 71         & 15    & 31    & 2     & 58         \\
			3                                                                     & Voice(VO)                                                                  & 3     & 7     & 2     & 58         & 7     & 15    & 2     & 58        \\
			\tblbottomrule 
	\end{tabulary}}\par 
\end{table*}

\bgroup
\fixFloatSize{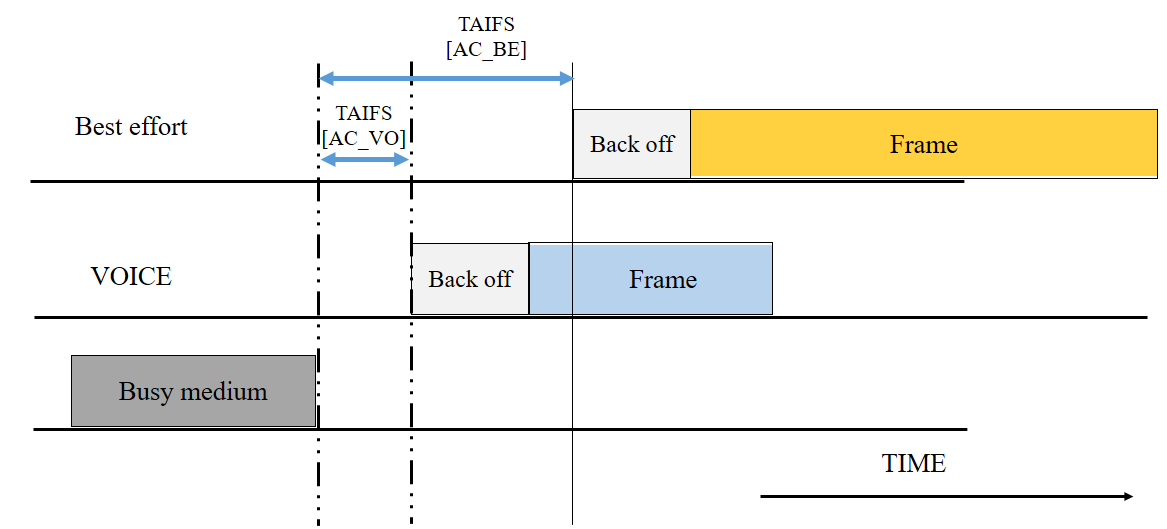}
\begin{figure*}[!htbp]
	\centering \makeatletter\IfFileExists{images/Figure2.png}{\includegraphics[width=.80\linewidth]{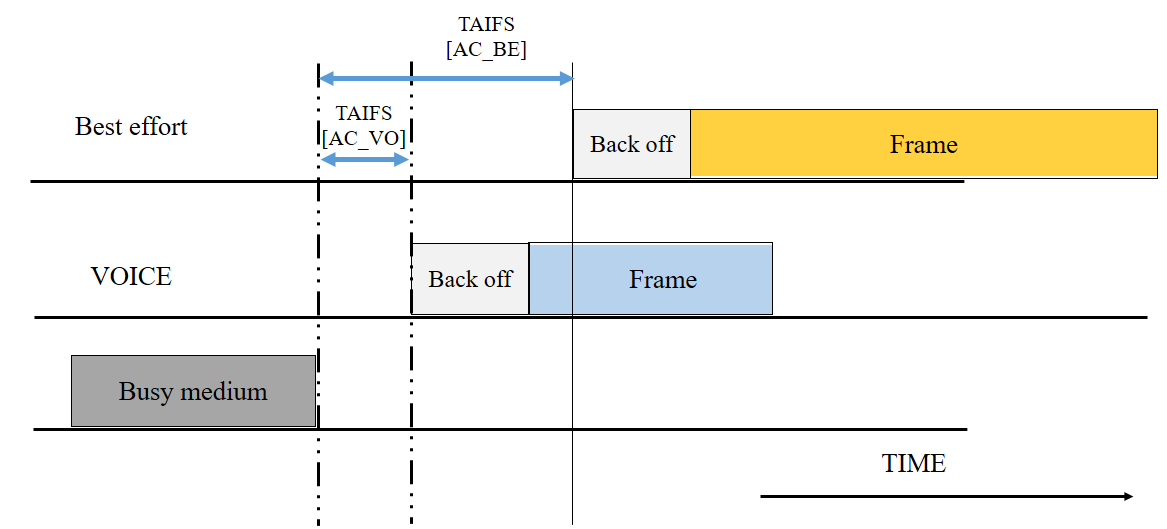}}{}
	\makeatother 
	\caption{{Example of contending for the medium access in EDCA IEEE 802.11p.}}
	\label{figure-b0dda45e191a46e975d0b852637b85d7}
\end{figure*}
\egroup

\subsection{ Overview of the H265/HEVC encoding modes }HEVC, like its predecessor H264/AVC, follows a hybrid video coding scheme. Both video coding standards have a two-layered high-level design consisting of a network abstraction layer (NAL) and video coding layer (VCL). The VCL includes all low-level signal processing, including inter- and intra-picture prediction, block partitioning, transform coding, in-loop filtering, and entropy coding. At the top-level, an HEVC sequence consists in a series of network adaptation layer (NAL) Units or NALUs. These NALUs encapsulate compressed payload data and include parameter sets containing key parameters used by the decoder to correctly decode the video data slices, which are coded video frames or parts of video frames \unskip~\cite{259927:5821226}.

More generally the gain brought by the HEVC makes it possible to envisage the transmission of video and particularly real-time video in circumstances presenting severe transmission conditions such as low bandwidth networks or with a high packet loss rate. Indeed, the video transmission in the VANET, which is judged a rather hostile network, requires compressing the information as much as possible and to have a good resilience to protect against possible losses. Psannis et al. \unskip~\cite{259927:5821225} demonstrated that the HEVC significantly outperforms its predecessors in terms of reducing temporal error propagation for variable wireless video environment applications. In their work, the application of an HEVC encoding pattern with a LD configuration was compared under different packet loss rates to conventional H.264/AVC, MPEG-4 part 2 and H.263 coding standards.

To guarantee operations with low latency, whether at the encoder or the decoder, the prediction from future images is forbidden. In AVC/H.264, the low delay constraint is generally satisfied using only P-images, losing the compression efficiency of directional motion estimation. In HEVC, a new type of B-picture, the generalized P and B (GPB) picture, has been introduced to offer low delay operations while offering a great coding performance \unskip~\cite{kim_mc_2014}. A GPB is a bi-predictive frame, which uses only past reference images for inter prediction.

HEVC offers several configuration modes \unskip~\cite{sze_high_2014}, depending on the intended application, in terms of coding efficiency, computational complexity, processing time and error resilience techniques. The two main encoding complexity configurations are:
\begin{itemize}
\item \relax the "high efficiency" mode that provides high-efficiency coding with a significant computational cost,
\item \relax the "low complexity" mode that offers reasonably high efficiency with low coder complexity. 
\end{itemize}
Moreover, HEVC proposes different structures of temporal prediction\unskip~\cite{bossen_hevc_2012}.
The first one is the All Intra (AI) configuration where all images are encoded as an Instantaneous Decoder Refresh (IDR) picture. This means that no temporal reference pictures are used to guarantee independence between the frames. The disadvantage of this configuration is that it produces high data rates, which is prohibitive for real-time transfer in networks with severe transmission conditions. This configuration is most preferred for real-time applications where the image is captured from the camcorder and instantly coded, then decoded at the destination.

\bgroup
\fixFloatSize{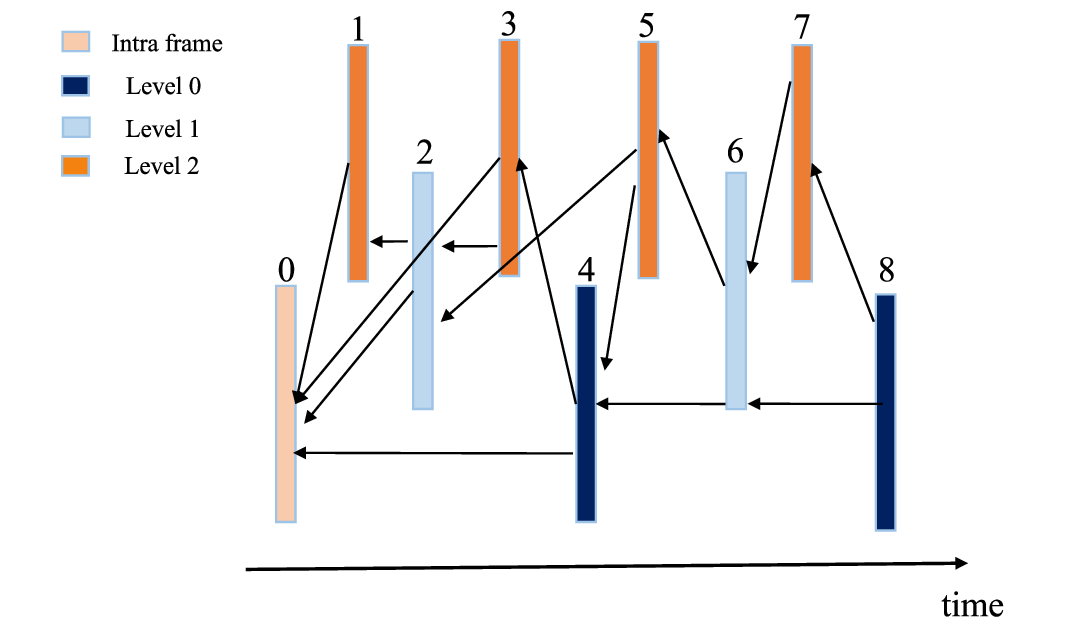}
\begin{figure*}[!htbp]
	\centering \makeatletter\IfFileExists{images/9.eps}{\includegraphics[width=.65\linewidth]{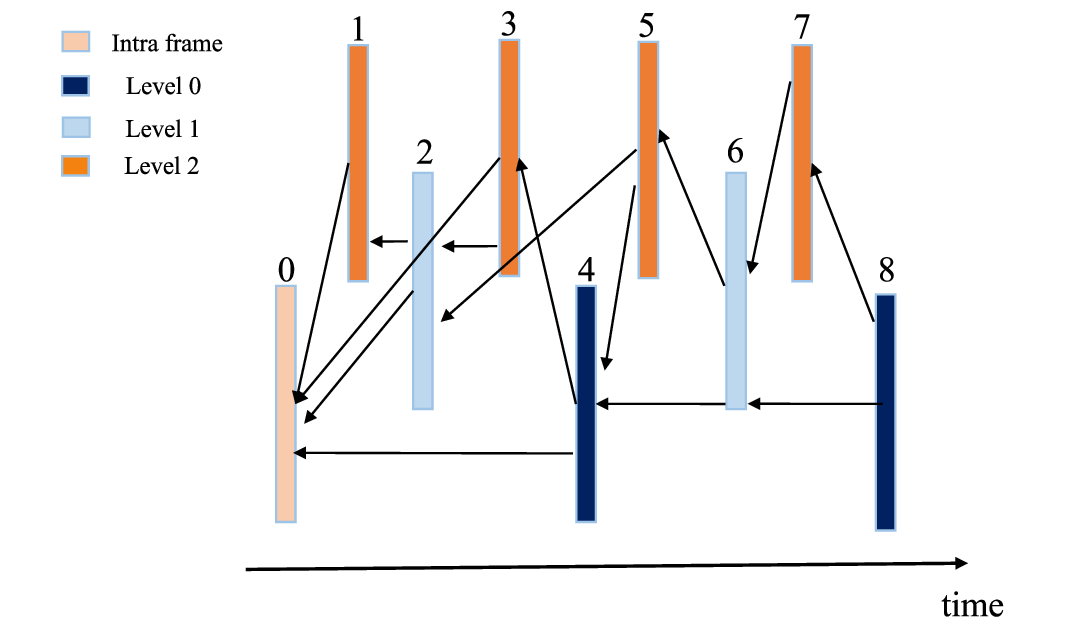}}{}
	\makeatother 
	\caption{{Graphical presentation of the Low Delay B configuration.}}
	\label{figure-1750b7e9373b3d28131b88d56f5457a2}
\end{figure*}
\egroup

The second temporal prediction structure of HEVC is the LD configuration where two coding configurations have been defined, called "low delay P" and "low delay B". In these conditions, only the first frame of a video clip is encoded as an IDR frame. The following frames are each encoded in a GPB-frame or as a P-frame for the low-delay P mode. For both modes, the P or GBP slices can only reference frames preceding the current frame in the display order. 

\mbox{}\protect\newline The frame coding and decoding orders are the same as the display order. This configuration is usually proposed for real-time applications systems where delay is a key parameter in QoS requirement, with a bit rate lower than the AI one. A graphical presentation of the LD-B configuration is depicted in Figure~\ref{figure-1750b7e9373b3d28131b88d56f5457a2} where the level corresponds to the temporal layer of each frame. 

\bgroup
\fixFloatSize{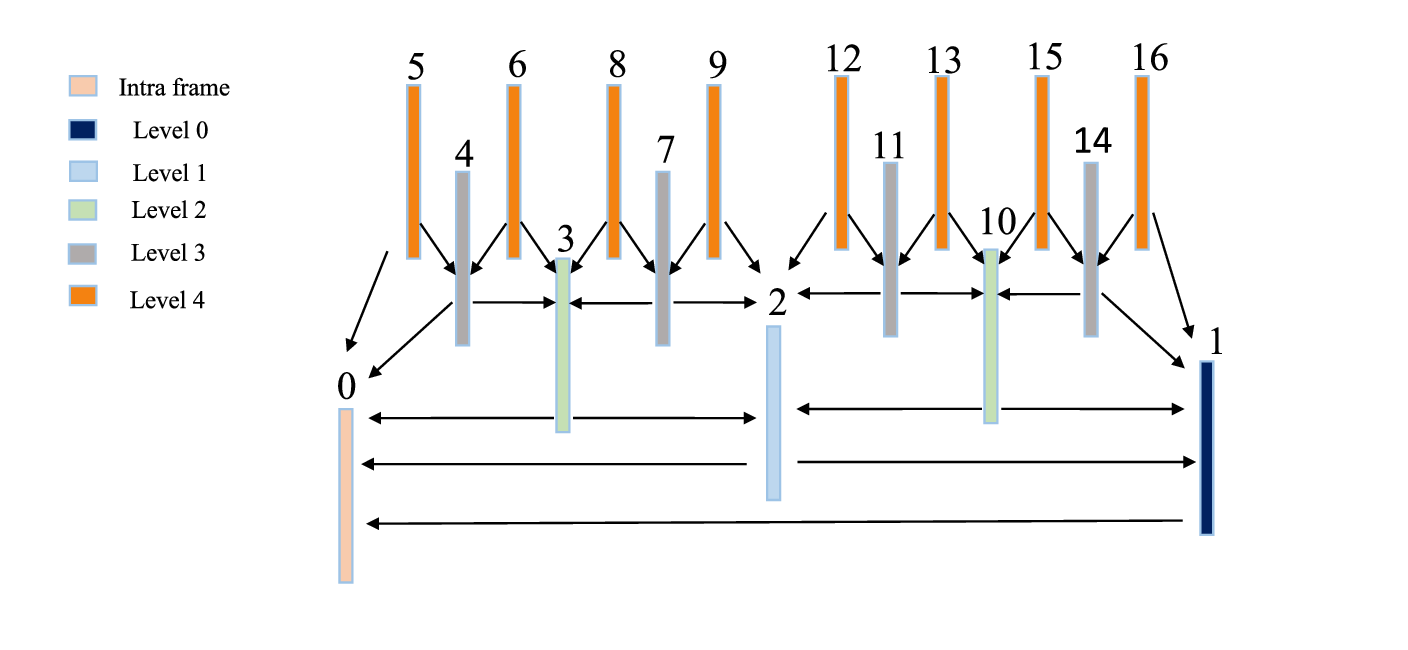}
\begin{figure*}[!htbp]
	\centering \makeatletter\IfFileExists{images/14.eps}{\includegraphics[width=.85\linewidth]{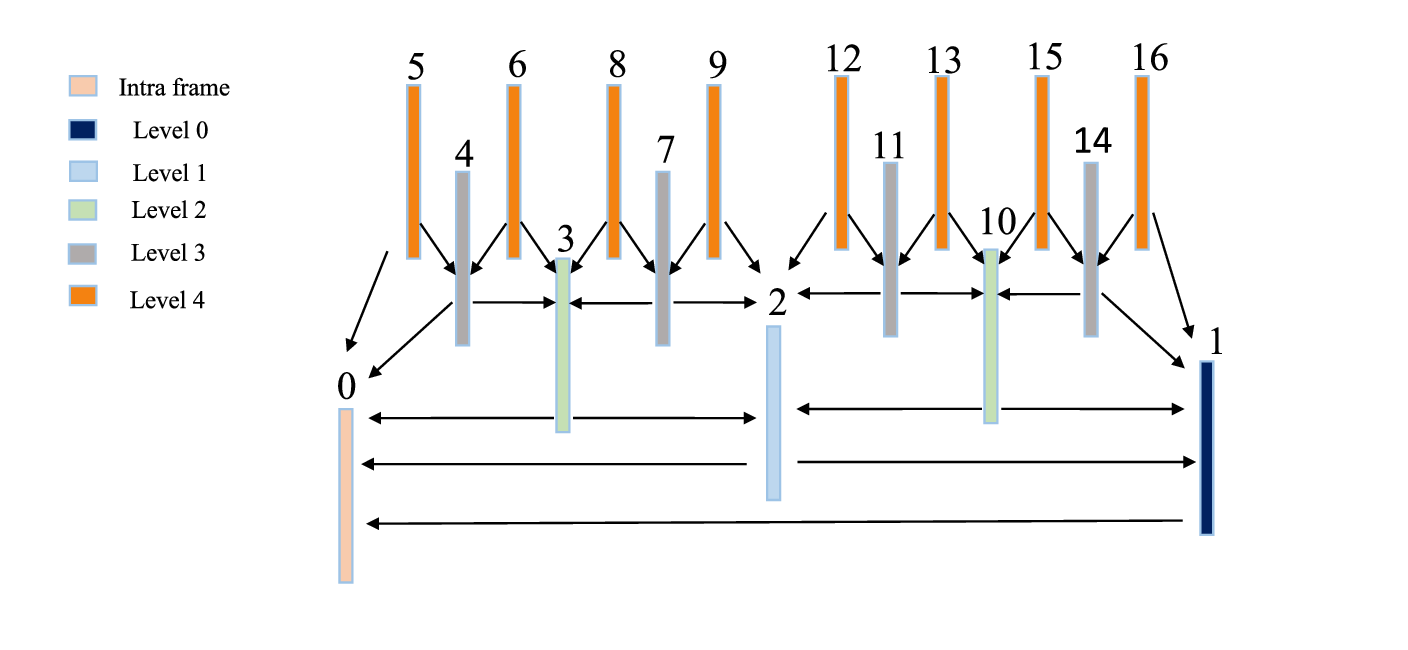}}{}
	\makeatother 
	\caption{{Graphical presentation of Random Access configuration.}}
	\label{figure-62a5bea82383b6c8f31e30b0fe5c5a9c}
\end{figure*}
\egroup

The temporal prediction of the Random Access (RA) configuration corresponds to the third type of temporal prediction in HEVC, as illustrated in Figure~\ref{figure-62a5bea82383b6c8f31e30b0fe5c5a9c}. It is understood that the encoding time and the decoding time of the frames are different from their display order. The frames are separated into a GoP. The first intra-frame of a video sequence is encoded as an IDR picture and the other intra pictures are encoded as intra non-IDR ("Open GoP") images. The frames located between successive intra pictures in the display order are coded as B-pictures. The GPB pictures can refer to the intra or inter-picture frames for the inter-prediction. The major advantage of this configuration is that it requires less encoding time than other inter-prediction configurations. This configuration is intended for applications requiring high compression capacity but no low latency constraint. Indeed, the delay time for this configuration depends on the coding process \unskip~\cite{259927:5821224}.

\subsection{Description of the proposed cross-layer approach}In this subsection, we describe our original cross-layer system. As stated above, the transmission of the video at the MAC layer of the conventional IEEE 802.11p standard is done only on the dedicated video AC. We propose here to exploit the other two ACs of lower priority in order to avoid network congestion and subsequent overflow drop of video packets. However, our system being developed and valid for different configurations, we nonetheless focus on a low delay aspect.
For this, we are considering three low complexity solutions for a low latency video transmission:
\begin{itemize}
  \item \relax Temporal prediction adaptation of the HEVC low delay encoding.
  \item \relax Static cross-layer mapping algorithm based on hierarchical HEVC encoding.
  \item \relax Adaptive cross-layer mapping algorithm based on hierarchical HEVC encoding.
\end{itemize}
 Moreover, the cross-layer system proposed exploits the hierarchization used by the HEVC for a better video packets mapping at the MAC layer level of IEEE 802.11p standard. The two proposed cross-layer mapping algorithms divide the different video packets into three different layers; this layerization is established according to the video structure:

For the low delay configuration: 

\begin{itemize}
  \item \relax Layer-1: includes the I-frames and the level-0 frames
  \item \relax Layer-2: includes the level-1 frames.
  \item \relax Layer-3: includes the level-2 frames. 
\end{itemize}

For the random access configuration: 

\begin{itemize}
  \item \relax Layer-1: includes the I-frames, level-0 and level-1 frames.
  \item \relax Layer-2: includes level-2 and level-3 frames.
  \item \relax Layer-3: includes the level-4 frames. 
\end{itemize}
  The choice of the frames distribution has been established based on the importance and the compressed data size of each frame. For the particular case of the All Intra configuration, no classification has been adopted.

\subsubsection{HEVC low delay encoding}The low delay configuration requires optimization in order to improve the error resilience and to reduce temporal error propagation in error-prone video transmission channels. The modification of the temporal prediction through the introduction of an Intra Random Access Point (IRAP) allows to stop error propagation in a video sequence. 

The HEVC standard includes three types of IRAP. In addition to known H.264/AVC (IDR) images, HEVC also supports clean random access (CRA) images and broken link access (BLA) images\unskip~\cite{259927:5821226}. A CRA image allows a master image to reference an image that precedes the CRA image in the display order and the decoding order. BLA images are used to indicate the splice points in the bit stream. 

\bgroup
\fixFloatSize{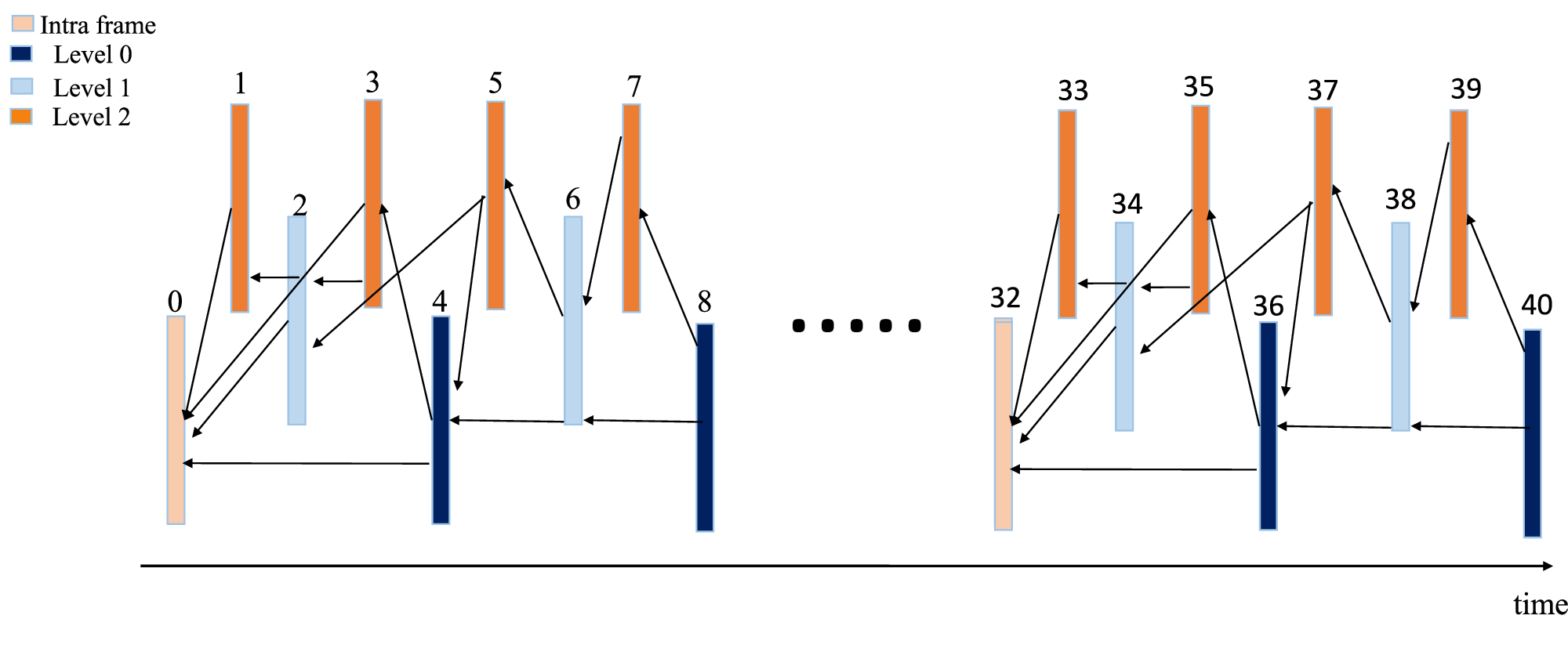}
\begin{figure*}[!htbp]
	\centering \makeatletter\IfFileExists{images/11.eps}{\includegraphics[width=1.0\linewidth]{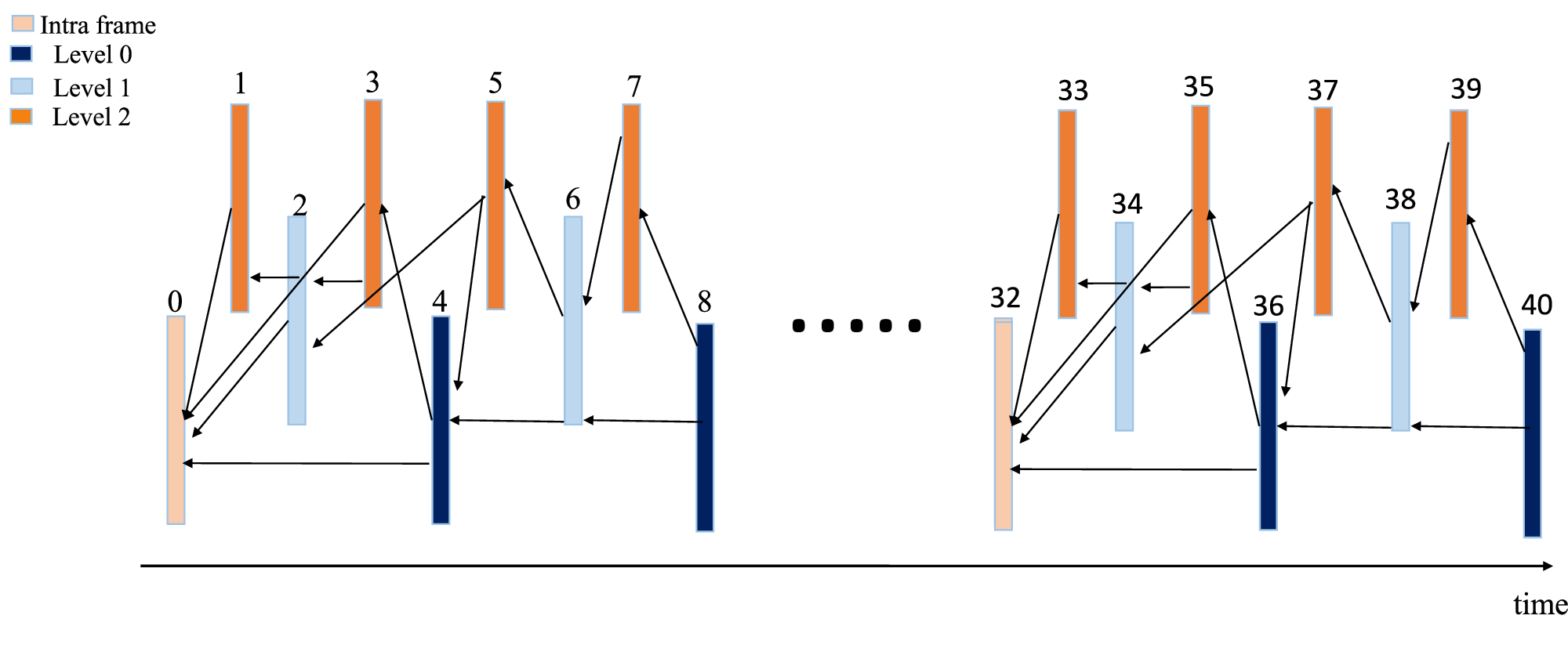}}{}
	\makeatother 
	\caption{{Graphical presentation of modified low delay configuration.}}
	\label{figure-b439cea2a1c953b9a02e8c36f5eef0f5}
\end{figure*}
\egroup

For our case and after a comparative study (section 4.1), we decide to introduce IDR images inserted at regular intervals. This completely refreshes the decoding process and launches a new coded video sequence (CVS). This means that neither the IDR image nor any image that follows the IDR image in the decoding order can depend on an image that precedes the IDR image in the decoding order. The modified low delay temporal prediction is shown in Figure~\ref{figure-b439cea2a1c953b9a02e8c36f5eef0f5}. This allows to avoid severe error propagation at the expense of a bit rate increase of about 10\% according to our estimation on several video sequences.

\subsubsection{Static mapping algorithm}Conforming to the adopted classification that varies according to the video structure, the images corresponding to the layer-1 are the images having the most importance. Indeed, the images of layer-1 have a great influence and affect in a way all the rest of the GoP. In this regard, any loss or degradation that may affect them will affect the entire GoP. The amount of information contained in the layer-1 frames is also bigger. In view of this information, we firstly propose to set up a static algorithm for each video structure where the layer-1 frames are mapped in all cases on the usable AC with the highest priority, which is the AC video. Then, the layer-2 frames with the second greatest importance will be mapped to the second queue available: the best effort AC. However, we adopt here the principle used by \unskip~\cite{259927:5821238} for a layerization of the video. In our static algorithm, video packets corresponding to layer-3 will be towards the last queue, the background traffic AC. 

This implies that for the AI configuration, this will not change anything compared to the EDCA standard and all frames will be delivered to the video queue buffer. As against the other configurations, the mapping differs. Indeed, for the LD or RA configurations, the images corresponding to the layer-2 frame will be mapped on the best effort AC and the layer-3 frame on the AC background traffic as suggested by Ksentini et al. \unskip~\cite{259927:5821238}. Figure~\ref{figure-95b6f4867808d0ba5a5f2ab54bf6671e} illustrated our static mapping algorithm. 
\bgroup
\fixFloatSize{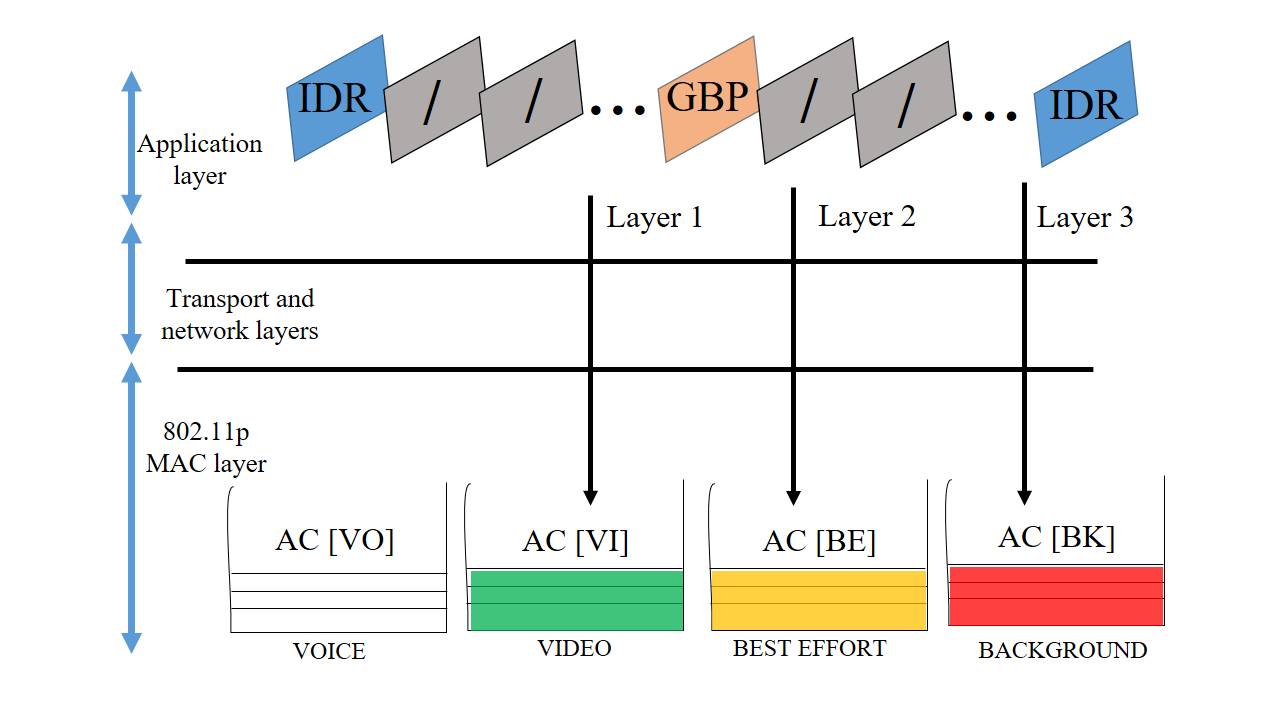}
\begin{figure*}[!htbp]
\centering \makeatletter\IfFileExists{images/10.png}{\includegraphics[width=.85\linewidth]{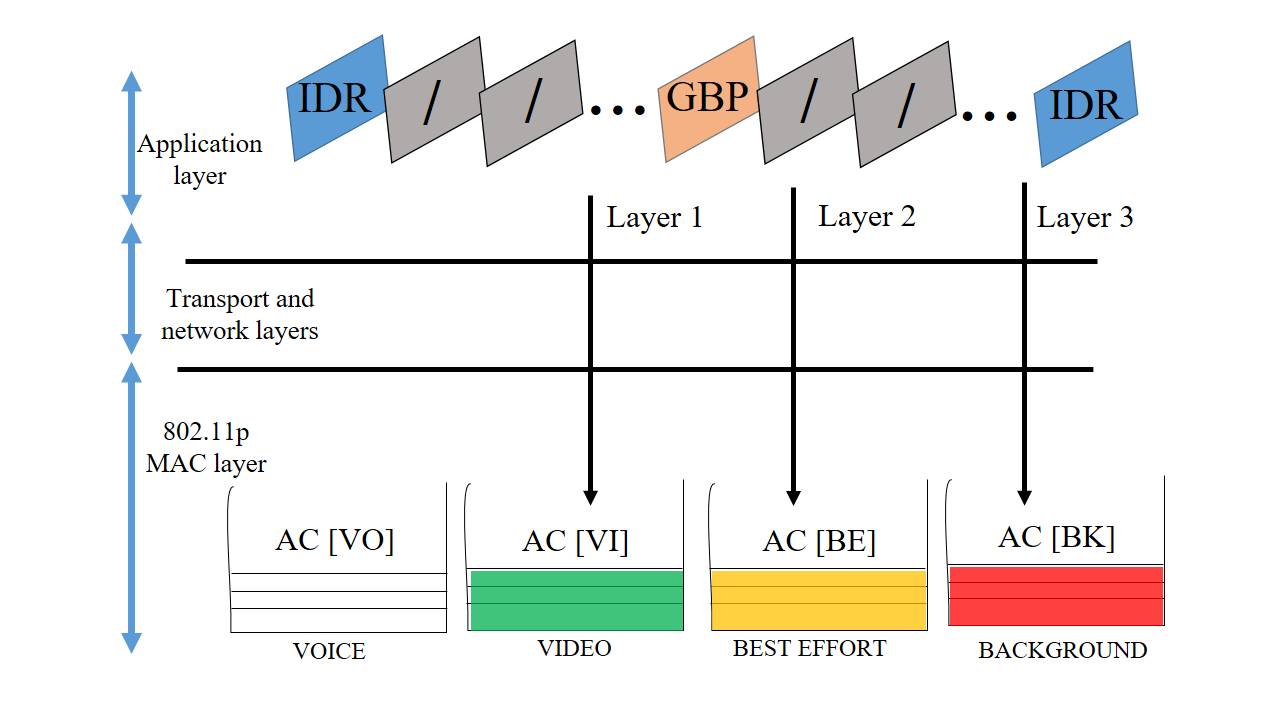}}{}
\makeatother 
\caption{{Illustration of the static cross-layer algorithm.}}
\label{figure-95b6f4867808d0ba5a5f2ab54bf6671e}
\end{figure*}
\egroup

\subsubsection{Adaptive mapping algorithm}The proposed adaptive mapping algorithm allocates dynamically to each packet video the most appropriated AC at the MAC layer. It considers the type of the temporal prediction structure of the video sequence, the importance of each frame and the state of the network traffic load. Moreover, still in order to benefit the most important images, we have to grant to each type of image a different mapping probability to ACs with lower priority, defined as\textit{ P}\ensuremath{_{\rm \_Layer}}. The probability is function of the importance of the frame which means $0\leq P_{\_Layer-1}\leq P_{\_Layer-2}\leq P_{\_Layer-3}\leq1. $

Otherwise, as previously stated the mapping also does depend on the state of the channel. The filling state of the AC queues reflects the network traffic load state. Indeed, the more the MAC queue buffer is filled, the more likely it is that the network overloaded. To manage and avoid network congestion we adopted two thresholds, qth\ensuremath{_{\rm high}} and qth\ensuremath{_{\rm low,}} the idea of this type of control comes from the principal of Random Early Detection (RED) mechanism. The adaptive mapping algorithm is based on the following expression originally exploited by Lin et al. in \unskip~\cite{259927:5821237}:
\begin{equation}
{P_{\_new}} = {P_{\_Layer}} \times \frac{{qlen\left( {AC\left[ {VI} \right]} \right) - qt{h_{low}}}}{{qt{h_{high}} - qt{h_{low}}}}\
\label{moneq2}
\end{equation}

Where\textit{ P}\ensuremath{_{\rm \_Layer}} is the initial probability which is based on the importance of the layer, \textit{qlen}(AC[\textit{VI}]) is the actual video queue length, qth\ensuremath{_{\rm high}} and qth\ensuremath{_{\rm low\ }}, are arbitrarily chosen thresholds which define the manner and degree of mapping to ACs with lower priority.\mbox{}\protect\newline Indeed, when \textit{qlen}(AC[VI]) is inferior to the qth\ensuremath{_{\rm low}}, all the packets are mapped in the AC[\textit{VI}]. When \textit{qlen}(AC[\textit{VI}]) is between qth\ensuremath{_{\rm low}} and qth\ensuremath{_{\rm high}}, \textit{P}\ensuremath{_{\rm \_new}} define the probability of packet mapped to AC[\textit{BE}]. When \textit{qlen}(AC[\textit{VI}]) is superior to the qth\ensuremath{_{\rm high}}, the video packets are mapped in the AC[\textit{BE}]. In this case, \textit{P}\ensuremath{_{\rm \_new}} defines the probability of the mapping towards the AC[\textit{BK}]. Figure~\ref{figure-8463842f7547dc666beb2fca2f6e5e26} illustrated our adaptive mapping algorithm. 
\bgroup
\fixFloatSize{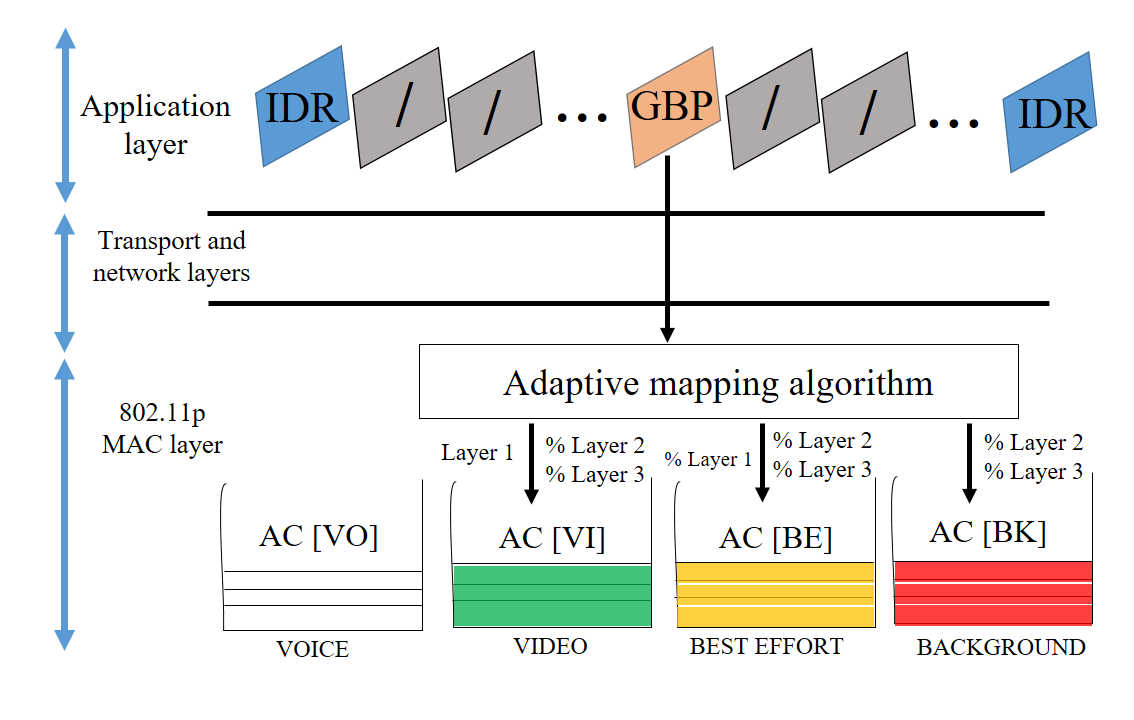}
\begin{figure*}[!htbp]
\centering \makeatletter\IfFileExists{images/1.png}{\includegraphics[width=.80\linewidth]{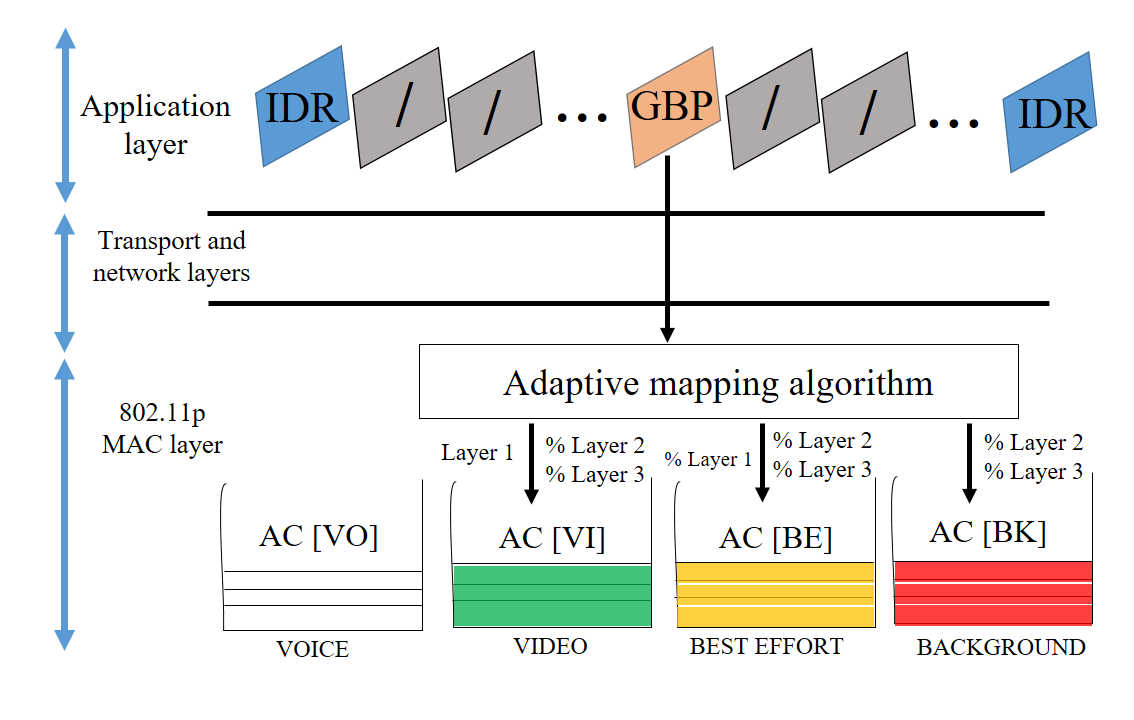}}{}
\makeatother 
\caption{{Illustration of the adaptive cross-layer algorithm.}}
\label{figure-8463842f7547dc666beb2fca2f6e5e26}
\end{figure*}
\egroup

\section{Framework and simulation set-up}
To evaluate the performances of the proposed solutions, we have implemented a realistic simulation of a video transmission in a vehicular framework. The adopted framework, illustrated in Figure~\ref{figure-f4b9e6f858bcb528137c2022f58c216f} is composed of three main blocks: a vehicular traffic simulator, a network simulator, and a video encoder/decoder. 

For the vehicular traffic simulator, we have chosen to use SUMO (Simulation of Urban Mobility) \unskip~\cite{259927:5821223}. This open-source simulator models the behavior of vehicles with urban mobility and considers vehicles interaction with each other, junctions, traffic lights, etc. Therefore, we use two different vehicular environments. For the first one, we have modeled a suburban traffic in the region of Valenciennes (France), with vehicles having random speeds and directions. For the second one, we modeled the highway between Paris and Lille (France) over 11 km with vehicles following the same road with the same speed. The road mobility, generated from SUMO, of the two environments studied, are used in the chosen network simulator NS2. For both cases, real maps were imported from OpenStreetMap (OSM) \unskip~\cite{259927:5821222}.

\bgroup
\fixFloatSize{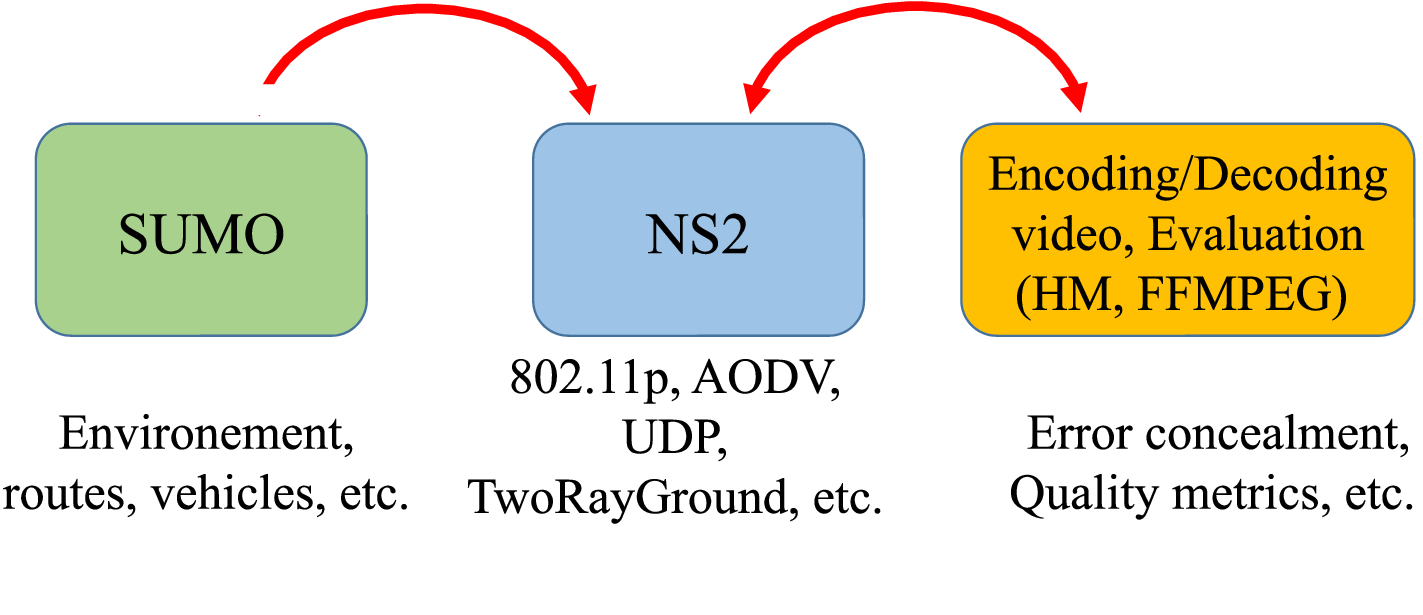}
\begin{figure*}[!htbp]
\centering \makeatletter\IfFileExists{images/4.eps}{\includegraphics[width=.70\linewidth]{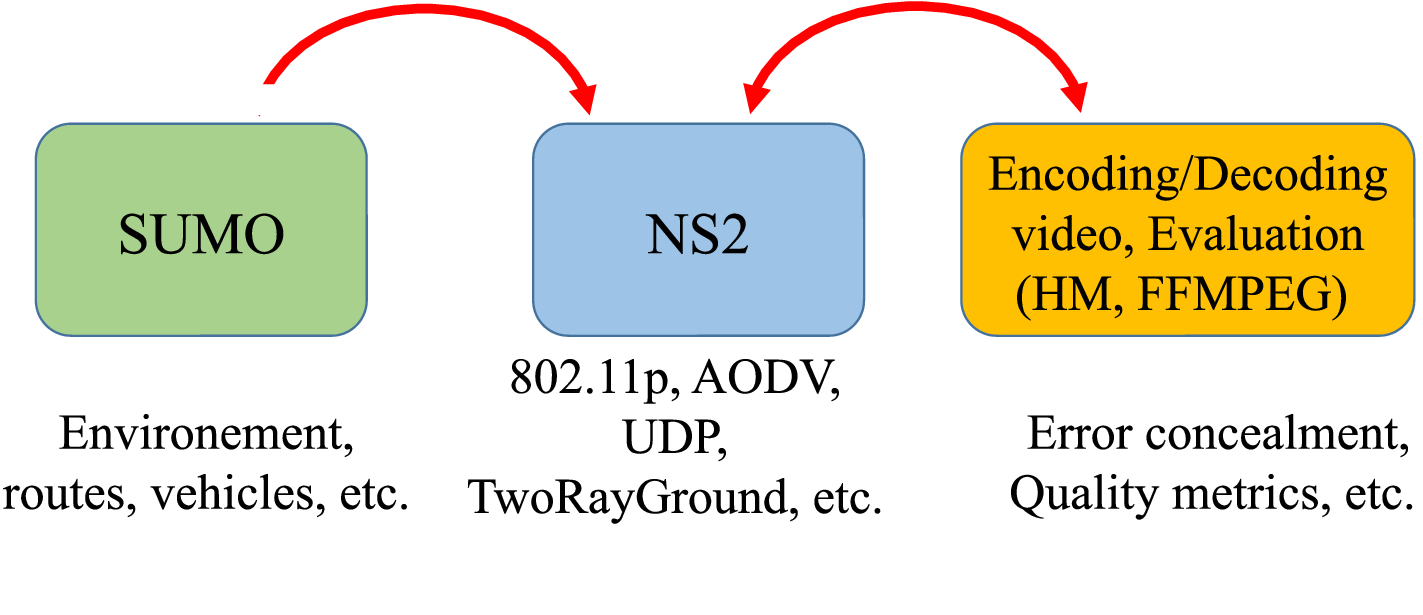}}{}
\makeatother 
\caption{{Representation of the VANET simulation scheme.}}
\label{figure-f4b9e6f858bcb528137c2022f58c216f}
\end{figure*}
\egroup
To allow the video transmission evaluation with such tools, we also have integrated the tool-set "Evalvid" \unskip~\cite{259927:5821221} in the framework. The videos meanwhile were encoded with the latest reference model HM (16.16) \unskip~\cite{259927:5821220}. For the choice of the appropriate sequences, the joint collaborative team on video coding (JCT-VC) introduced the notion of class in the choice of sequences to be processed. Indeed, test video sequences are classified into six classes (class A to F) \unskip~\cite{259927:5821226}. Class C is intended to evaluate the performance of set mobile video applications with an image resolution of 832 x 480 pixels. In addition to this, ITU-T Recommendation P.910 indicates that in the selection of test sequences, it may be useful to compare the relative spatial information and the temporal information found in the different available sequences \unskip~\cite{259927:5821219}. Generally, the difficulty of compression is directly related to the temporal information (TI) and the spatial information (SI) of a sequence. For our case, we decided to use the following four class C sequences: RaceHorses, PartyScene, BasketballDrill and BQMall. Figure~\ref{figure-70606d7d2e9cb0d657e65136f008254b} illustrates the SI \& TI of these different sequences. The four sequences have a duration of 10 seconds but different frame rate, respectively 30, 50, 50, and 60 fps. In regards to the encoding, the sequences were encoded with three HEVC configurations: LD, RA, and AI. Furthermore, LD and RA have been encoded with the same GoP size of 32 frames.

\bgroup
\fixFloatSize{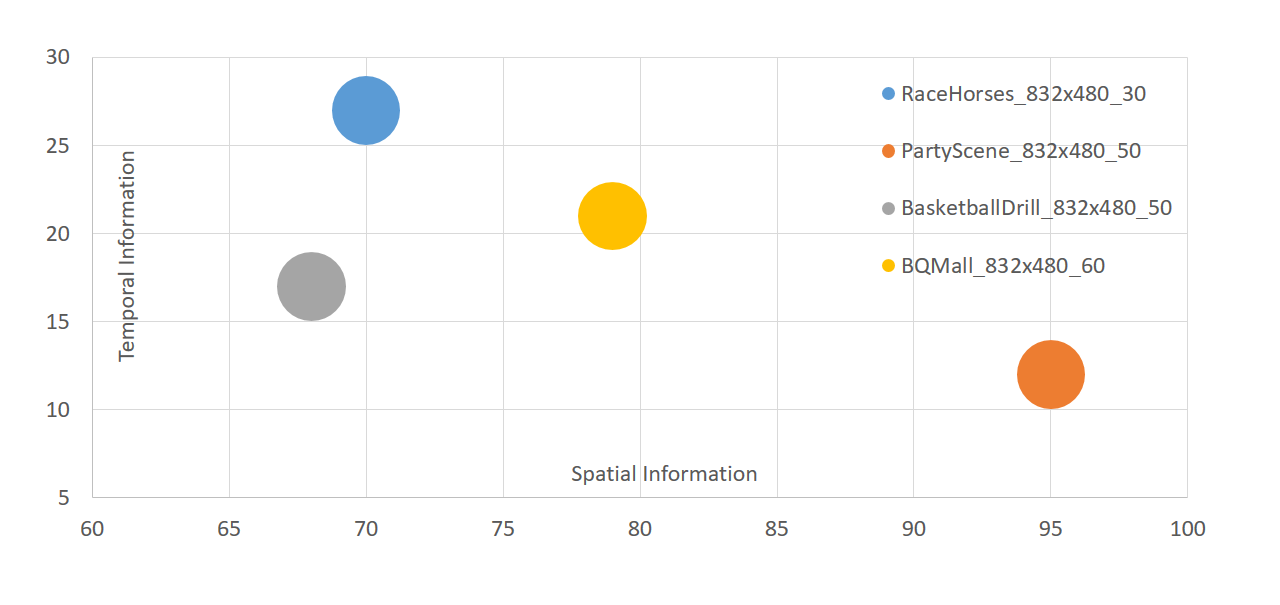}
\begin{figure*}[!htbp]
	\centering \makeatletter\IfFileExists{images/3.png}{\includegraphics[width=.75\linewidth]{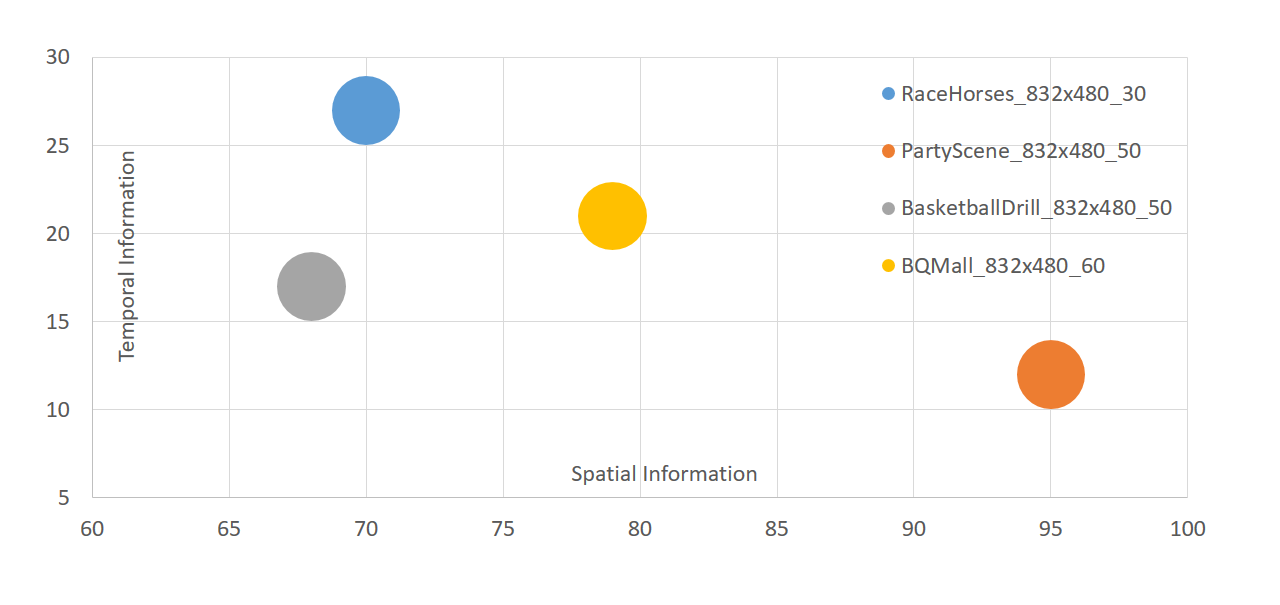}}{}
	\makeatother 
	\caption{{Representation of the SI and TI of test sequences used.}}
	\label{figure-70606d7d2e9cb0d657e65136f008254b}
\end{figure*}
\egroup
To ensure video decoding continuity, even in extreme cases with a high percentage of lost packets, the "Evalvid" tool-set allows the video to be transmitted and reconstructed at the receiver side. This tool-set for the evaluation of the video quality integrates an error concealment process based on last decoded frame copy \unskip~\cite{259927:5821221}. Radio wave propagation model influences the probability of receiving packets. Ray tracing simulation gives the most detailed view of the radio wave propagation. However, the execution time remains too important due to the required computation cost.\mbox{}\protect\newline Thus, the scientific community often applied simplified models to characterize the radio propagation. However, the literature differentiates between deterministic and probabilistic models, both of which characterize the attenuation of the strength of the radio signal over distance. As far as we are concerned, the radio propagation model used is TwoRayGround making it possible to give a realistic modeling representation of free space vehicular channel \unskip~\cite{4656896,6814735}.
\begin{table*}[!htbp]
	\caption{{ Simulation parameters of the VANET scenario.} }
	\label{table-wrap-7aca930dbd46ade54a9716bf30175c1a}
	\def\arraystretch{1}
	\ignorespaces 
	\centering 
	\begin{tabulary}{\linewidth}{LL}
		\hline 
		Parameter & Value\\
		\tblmidrule 
		Radio-propagation model &
		TwoRayGround\\
		Video play time &
		10s\\
		Maximum Transfer Unit (MTU) &
		1024 Bytes\\
		Routing protocol &
		AODV\\
		Transport protocol &
		UDP\\
		Maximum packet in interface queue (IFQ) &
		50 packets\\
		Used metrics &
		PSNR\mbox{}\protect\newline Packet delivery.\\
		Scenario &
		V2V\\
		\tblbottomrule 
	\end{tabulary}\par 
\end{table*}
\mbox{}\protect\newline Furthermore, as described before, the standard used is the IEEE 802.11p one. Which includes a MAC layer using CSMA/CA with QoS support and a PHY layer operated in the 5.9 GHz frequency band with OFDM modulation. As for the routing protocol, we chose the Ad Hoc On-Demand Distance Vector (AODV) protocol which is a reactive protocol and which through our own experiments or according to the literature \unskip~\cite{259927:5821217} shows a performance advantage compared to the other routing protocols. In regards to transport layer protocol and to guarantee a minimum latency, we have chosen to work with User Datagram Protocol (UDP). Table~\ref{table-wrap-7aca930dbd46ade54a9716bf30175c1a} summarizes the main parameters of the simulation. 

The parameters established for the system are as follows: \textit{P}\ensuremath{_{\rm \_Layer}}, is fixed at 0 for the layer-1 frame, 0.6 for the layer-2 frame and 0.8 for the layer-3 frame.\mbox{}\protect\newline As regards to the thresholds, they are set to 20 packets for qth\ensuremath{_{\rm \_low}} and 45 for \textit{qth}\ensuremath{_{\rm \_high}} knowing that the maximum packet number in the interface queue (IFQ) is 50.

\section{Simulation results}
Several experiments have been carried out to demonstrate the effectiveness of the proposed mechanisms. We have defined two distinct scenarios:
\begin{itemize}
	
 \item \relax The first scenario: only video is transmitted on the network. The goal here is to evaluate the quality of the video over time. Also, to demonstrate the behavior of the mechanisms through the packets mapping management and the state of filling the various ACs. Packet loss effect on video quality has also been addressed. In this scenario, "BQmall" video sequence was transmitted through the implemented suburban environment, it has been encoded at a rather high bit rate of 2.5 Mbps.

\item \relax The second scenario: video coexists with other types of streams. Indeed, we simulate with the video transmission, the transmission of a voice traffic in the AC [VO] but also a TCP stream in the AC [BE] and a UDP stream on the AC [BK]. As for the videos data used, we simulated the transmission of the four sequences previously mentioned in the highway environment. These videos have been encoded at 1.5 Mbps.
\end{itemize}

\subsection{HEVC Intra Random Access Point effect analysis}In order to study the effect on video received quality of the different Intra Random Access Points (IRAP) introduced by HEVC, we have simulated in the first scenario (suburban environment) the transmission of different video streams encoded with the different intra-coded images. Figure~\ref{figure-9d5b56dccfd234a092a0b7267c7e674d} illustrates the effect of different encoding types faced with the same channel conditions. The curves concern a "standard" LD configuration and three LD configurations with one intra-coded image every 32 frames (I-frame, CRA or IDR frame). We can see that the transmission of a video encoded with the "standard" LD configuration recommended by the HEVC standard causes a break at the reconstructed video sequence as soon as the first frame is lost. As a result, it is no longer possible to correctly decode the video stream. The IRAP image periodic insertion makes it possible to return to good PSNR scores if transmission allows it. CRA and I-frames allow referencing an image that precedes the IRAP image. This results in a stepwise flow recovery for I-frame or slower for CRA. Unlike IDR, which quickly and efficiently refers to good PSNR quality. The average PSNR of the different reconstructed video streams is given in Table~\ref{table-wrap-7e3ef05c6380df29d0061a9721cbb104}. We find that the standard structure takes longer to be affected by a loss and this is due to the smaller flow variation. This result can be explained by the absence of an intra-encoded image after the first image. However, this resulted in a better PSNR average than an I-frames insertion. We can also note that the average PSNR for the insertion of an IDR is better than the insertion of a CRA due to the harsh state of the channel and the numerous image losses. As the vehicular environment is rather severe, we decided to adopt an encoding with an IDR frame every 32 images.
\bgroup
\fixFloatSize{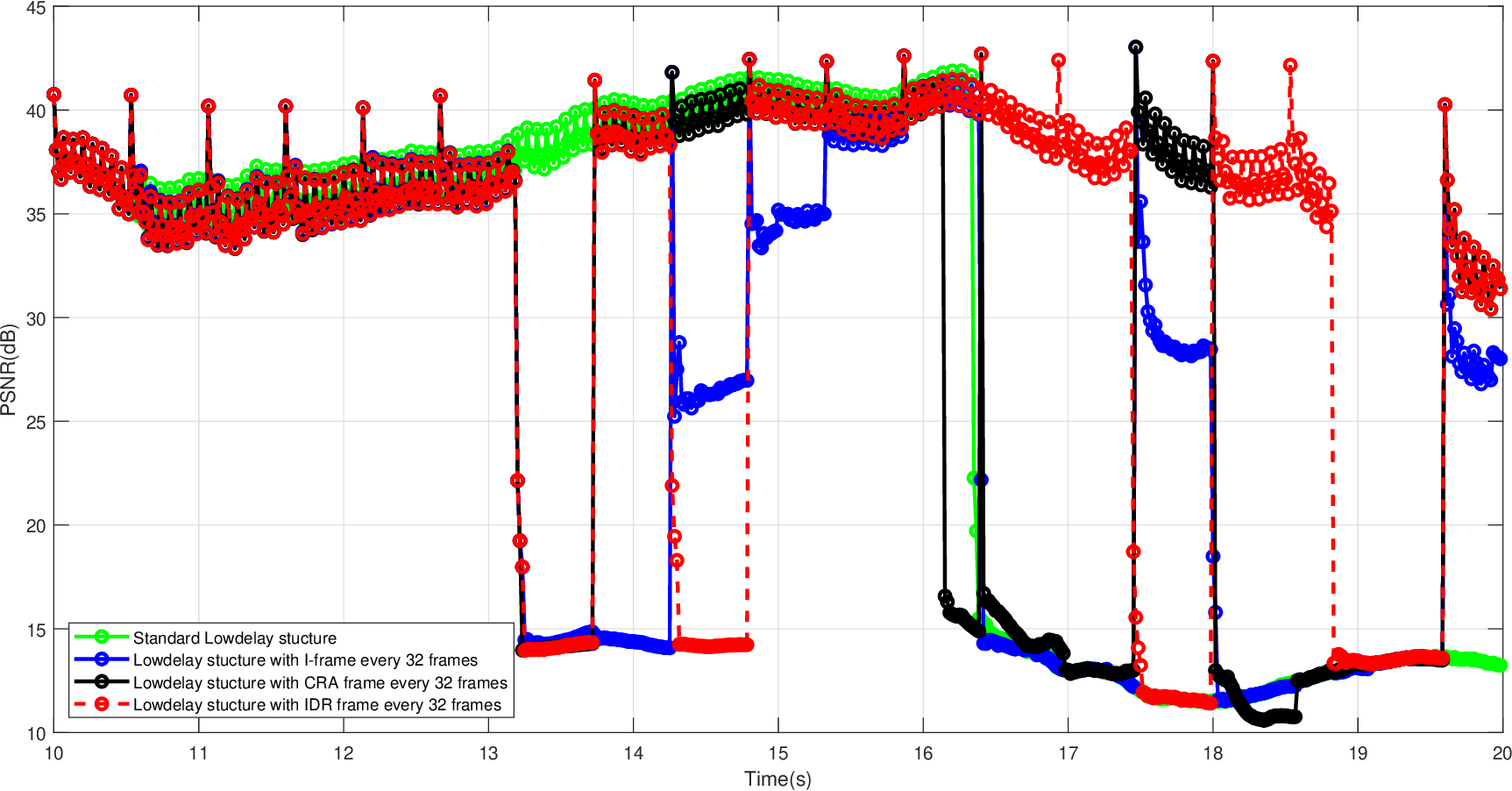}
\begin{figure*}[!htbp]
\centering \makeatletter\IfFileExists{images/12.eps}{\includegraphics[width=1.0\linewidth]{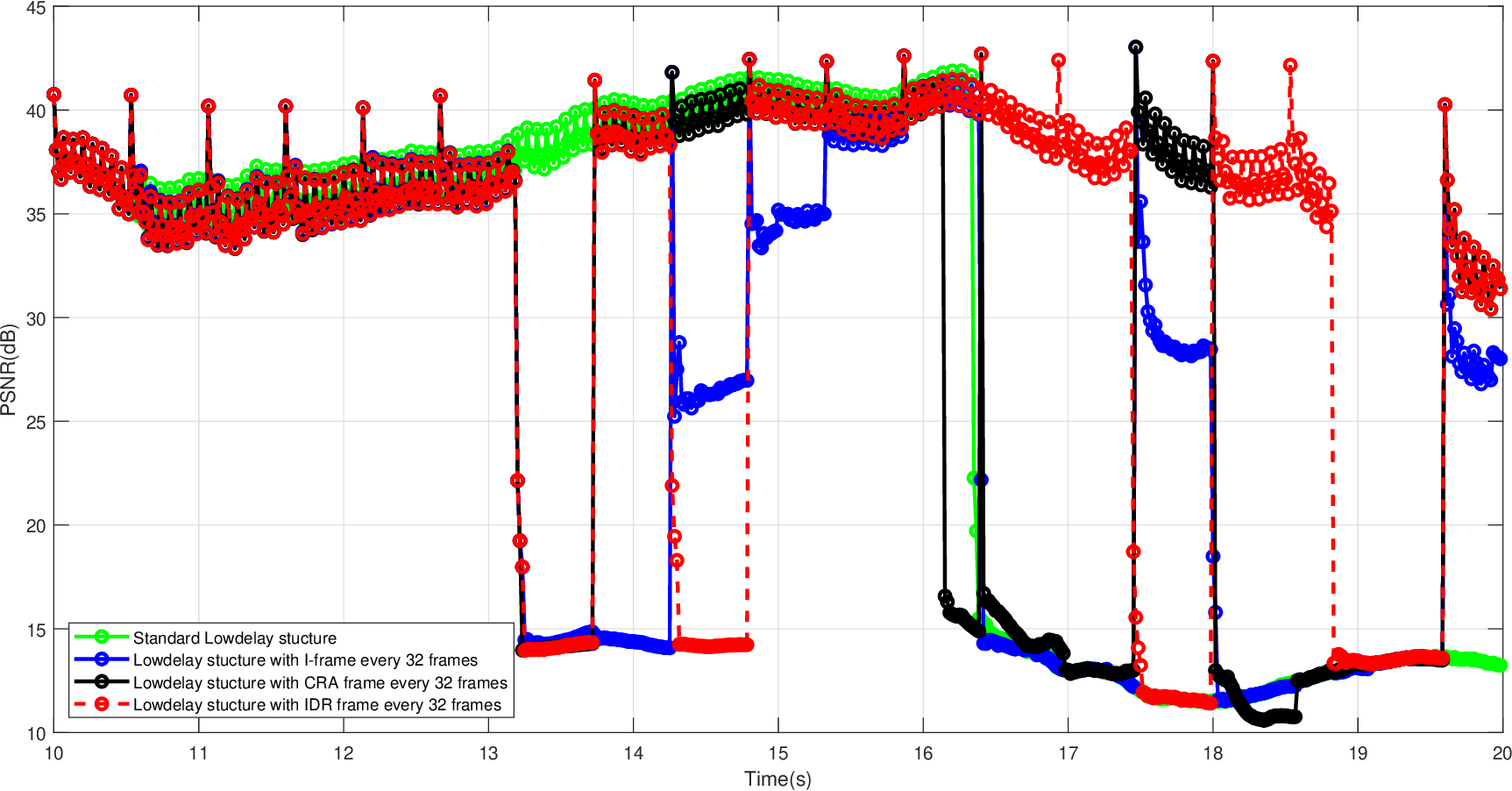}}{}
\makeatother 
\caption{{The PSNR temporal evolution for the different encoding schemes.}}
\label{figure-9d5b56dccfd234a092a0b7267c7e674d}
\end{figure*}
\egroup

\begin{table*}[!htbp]
\caption{{ Average PSNR results.} }
\label{table-wrap-7e3ef05c6380df29d0061a9721cbb104}
\def\arraystretch{1}
\ignorespaces 
\centering 
\begin{tabulary}{\linewidth}{LLLLL}
\hline 
 & Low delay structure & I-frames & CRA & IDR\\
\tblmidrule 
Average PSNR (dB) &
  28.90 &
  26.85 &
  29.21 &
  31.71\\
\tblbottomrule 
\end{tabulary}\par 
\end{table*}

\subsection{Evaluation of the video quality with the different mapping algorithms}We consider again here the first scenario. The transmission of the video contents with the three different mapping methods described in Section 3 allows evaluating the gain brought by the adaptive method as illustrated in Figure~\ref{figure-e3799303195a5ba5a320f6ba152afaab}. The curves represent the PSNR evolution of the three mapping algorithms. It is clear that the adaptive method (in red) presents a better quality in terms of PSNR. The static method has good PSNR scores on some peaks representing good reception of IDR images. Nevertheless, the rest of the GoP frames have a bad PSNR score, consequence of the loss of the image packets. Anyway, the video quality is still better than the EDCA method. This last method exhibits a poor PSNR quality for many frames sequence. The Figure~\ref{figure-e3799303195a5ba5a320f6ba152afaab} also allows to see the intra-frame reference image loss effect in a GoP. Indeed, first frame loss in a GoP leads to a low PSNR for all the rest of the GoP. The zoomed part allows us to visualize the previously discussed states and will serve for a deeper analysis thereafter.

\bgroup
\fixFloatSize{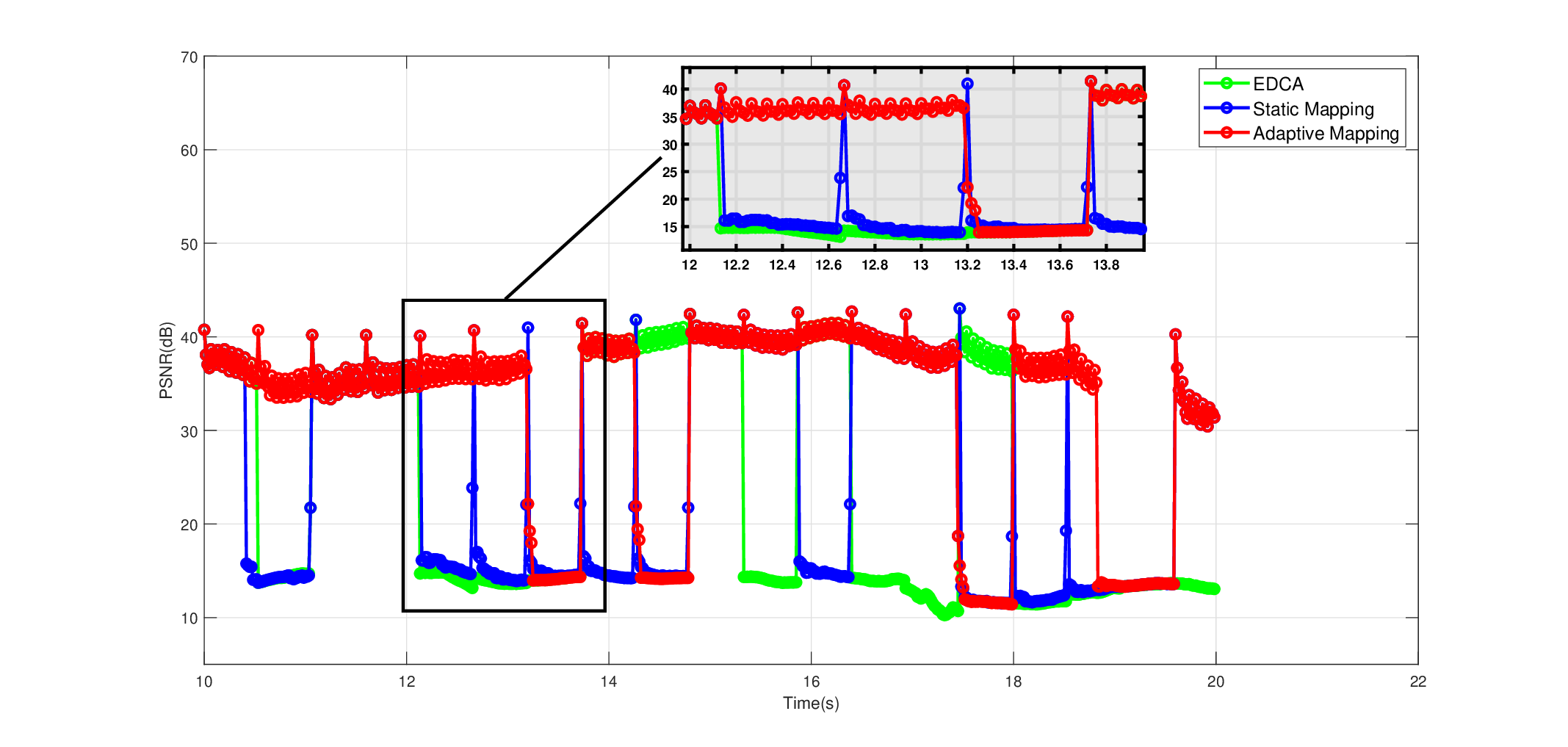}
\begin{figure*}[!htbp]
%\centering
\hspace{-2.0cm}
\makeatletter\IfFileExists{images/8.eps}{\includegraphics[width=1.20\linewidth]{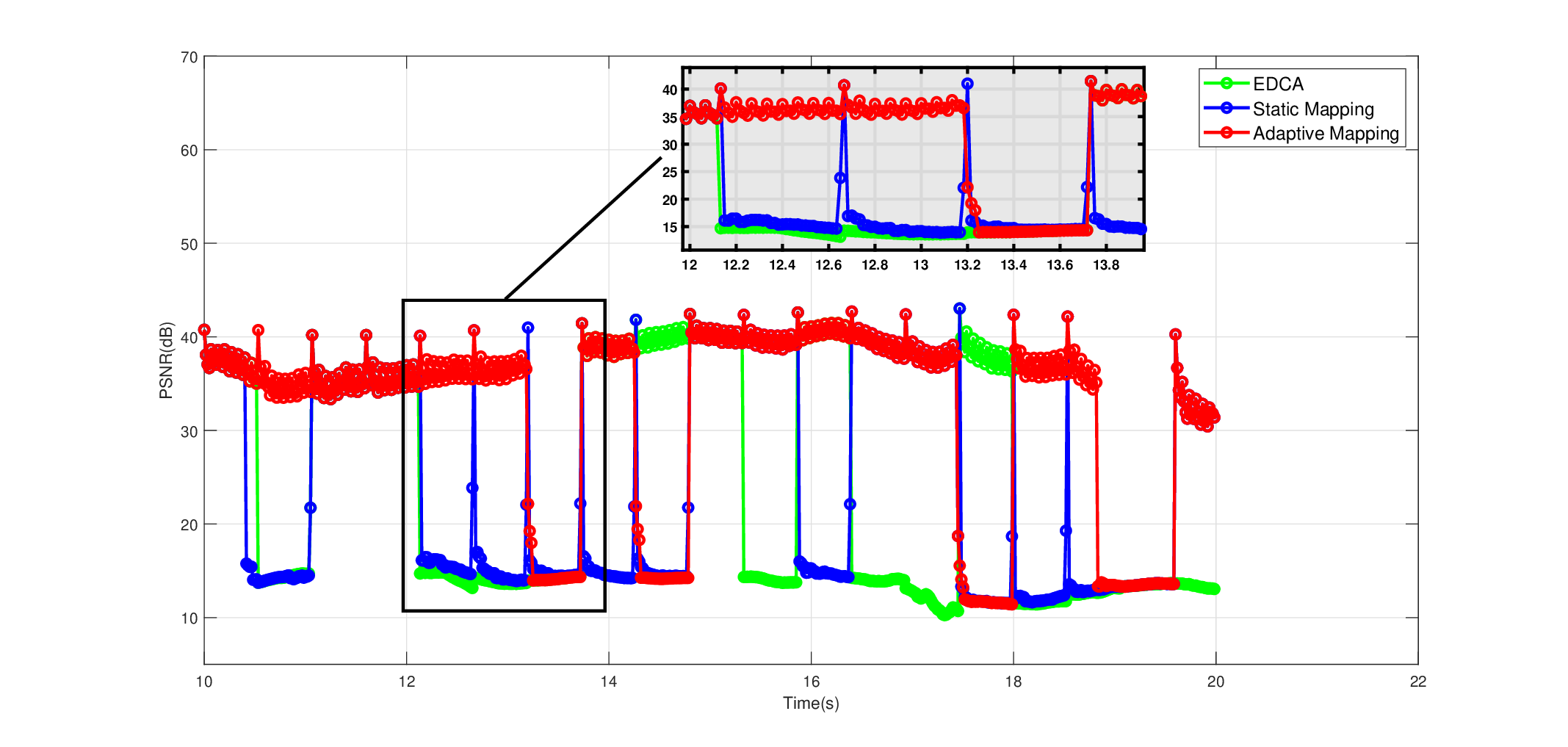}}{}
\makeatother 
\caption{{The variation of the PSNR for the different mapping algorithms: EDCA (green), static (blue) and adaptive (red); the letterbox shows an enlarged part of the curves for a 2s duration.}}
\label{figure-e3799303195a5ba5a320f6ba152afaab}
\end{figure*}
\egroup
The video packets classification as well as the exploitation of the resources provided by the IEEE 802.11p standard allows to limit packet losses and specially to protect the most important video packets. In Table~\ref{table-wrap-6ee8892e4ee35675e4b4101fa41eefaa}, we verify that the total number of lost packets significantly decreases as a function of the method efficiency. This is particularly true for the packets belonging to prioritized frames. The EDCA method has a packet loss number of 75 while it is reduced to 11 for the adaptive method. In detail, the packet loss is balanced between the different layers for the EDCA. For both the static and adaptive methods, the imbalance also depends on the importance of the layer. The protection of layer-1 frames makes it possible to go from 19 lost packets for the EDCA to 9 for the static algorithm and only 2 for the adaptive algorithm. However, the adaptive method manages to better protect the packets of the most important layers. This translates into better video quality as shown by the average video sequence PSNR.
\begin{table*}[!htbp]
\caption{{ Average PSNR and number of packets lost for each mapping algorithm} }
\label{table-wrap-6ee8892e4ee35675e4b4101fa41eefaa}
\def\arraystretch{1}
\ignorespaces 
\centering 
\begin{tabulary}{\linewidth}{LLLLLL}
\hline 
\multirow{2}{*}{\begin{tabular}[c]{@{}c@{}}Mapping \\ Algorithm\end{tabular}} & \multirow{2}{*}{\begin{tabular}[c]{@{}c@{}}Average\\ PSNR (dB)\end{tabular}} & \multicolumn{4}{c}{Number of packet lost}\\
 &
   &
  Layer-1 frame &
  Layer-2 frame &
  Layer-3 frame &
  Total\\
  \tblmidrule 
EDCA &
  23.86 &
  19 &
  19 &
  37 &
  75\\
Static mapping &
  24.42 &
  9 &
  11 &
  19 &
  39\\
Adaptive mapping &
  31.71 &
  2 &
  3 &
  6 &
  11\\
\tblbottomrule 
\end{tabulary}\par 
\end{table*}

\subsection{Analysis of the filling status of the queues for the different mapping algorithms}In order to explain the gain in quality, we analyze the queues filling for the three mapping methods. The analysis is done for the duration time which corresponds to the enlarged part of Figure~\ref{figure-e3799303195a5ba5a320f6ba152afaab}. Figure~\ref{figure-574ce0ae2b9c651fda30cf3f1828c5fd} to 14 shows the evolution over time of the following three parameters for the EDCA, static and adaptive algorithms, respectively:

\begin{itemize}
	\item \relax the queuing status of the VI, BE and BK queues, 
	\item \relax the reconstructed PSNR video quality, 
	\item \relax an indication of lost frames for the three layers
\end{itemize}
The left Y, right Y and X (abscissa) axes respectively represent the queue length, the video PSNR and the simulation time.

For the EDCA algorithm, Figure~\ref{figure-574ce0ae2b9c651fda30cf3f1828c5fd}, only the AC video is used. This AC queue reaches three times, during the enlarged part, the queue maximum capacity that causes drops of subsequent packets. These cause image losses that occur only on layer-1 frames because of their larger sizes. In the details, the lost packets are those of the IDR frames. This affects all the GoP that depends on the IDR frame, which explains the low PSNR values during the three GoPs. Considering the following GoP, no overflow takes place and the PSNR indicates good quality.

\bgroup
\fixFloatSize{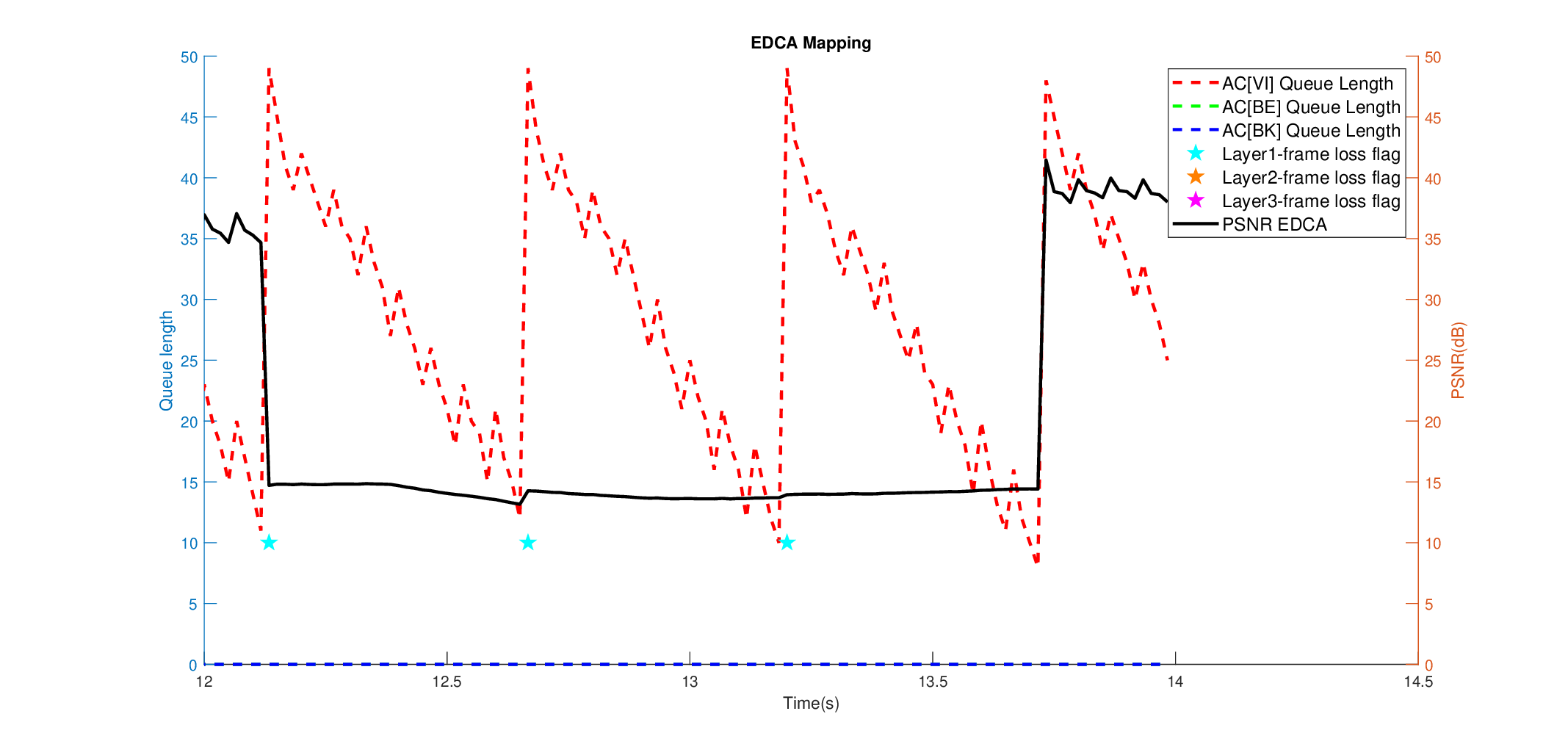}
\begin{figure*}[!htbp]
\centering \makeatletter\IfFileExists{images/2.eps}{\includegraphics[width=1.05\linewidth]{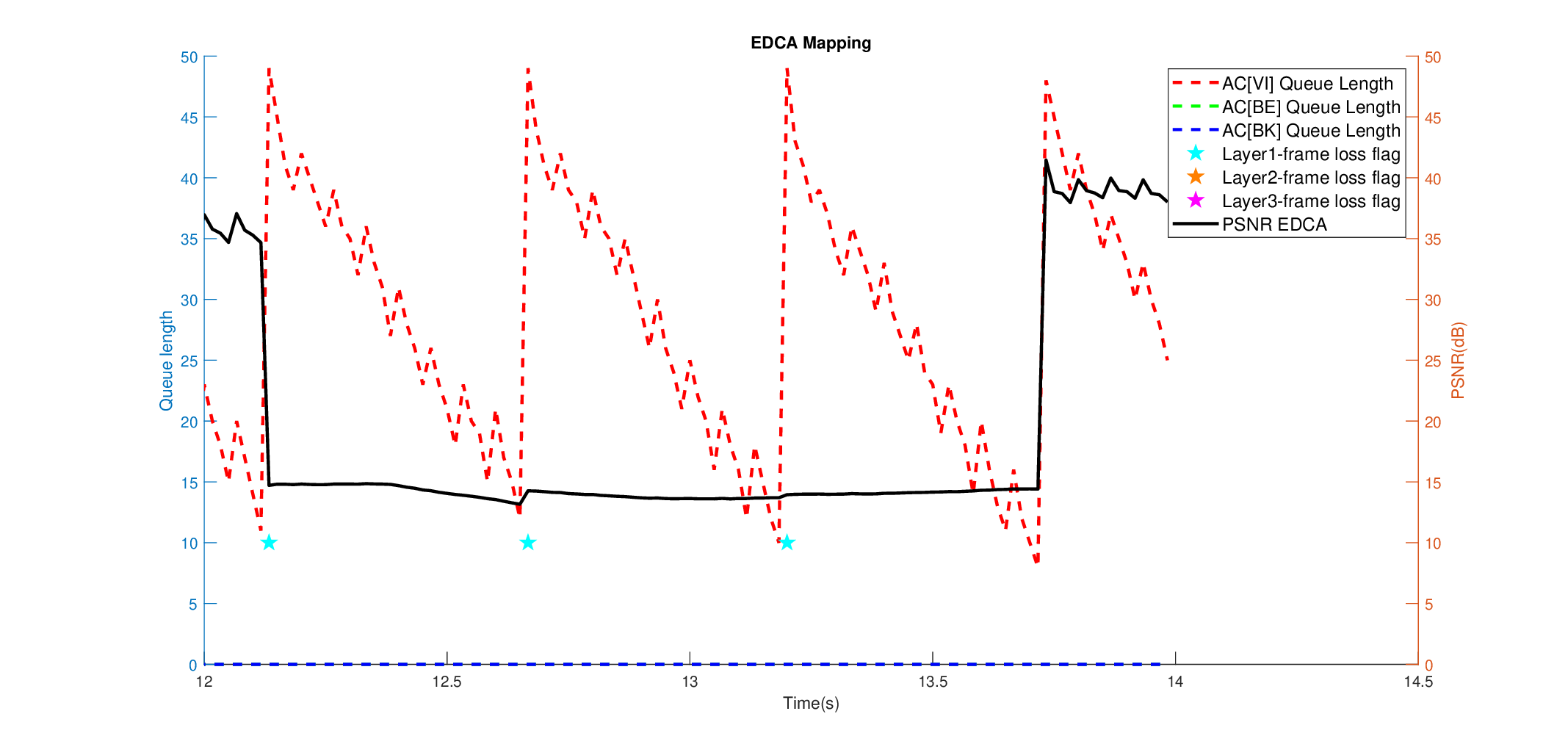}}{}
\makeatother 
\caption{{EDCA mapping algorithm: filling state of the queue, PSNR video quality and indication of lost images over time.}}
\label{figure-574ce0ae2b9c651fda30cf3f1828c5fd}
\end{figure*}
\egroup

Exploiting the two other queues of lower priority allows reducing frame packet drop number, especially for the most important frames. Figure~\ref{figure-ae03c2e7f761d33c007bf6285f67d8d7} relates the static algorithm, shows that the AC[VI] is not fully exploited and the AC[BE] is practically not at all. The AC [BK] which is the most disadvantaged is the most fulfilled. The filling of the three queues does not reach the maximum capacity on the zoomed part. Hence frame-packet drops are avoided. We can, however, observe losses in the layer-3 frames that are due to the arrival of video packets with significant delay. We can also found that layer-1 and layer-2 frames packets are correctly received.

\bgroup
\fixFloatSize{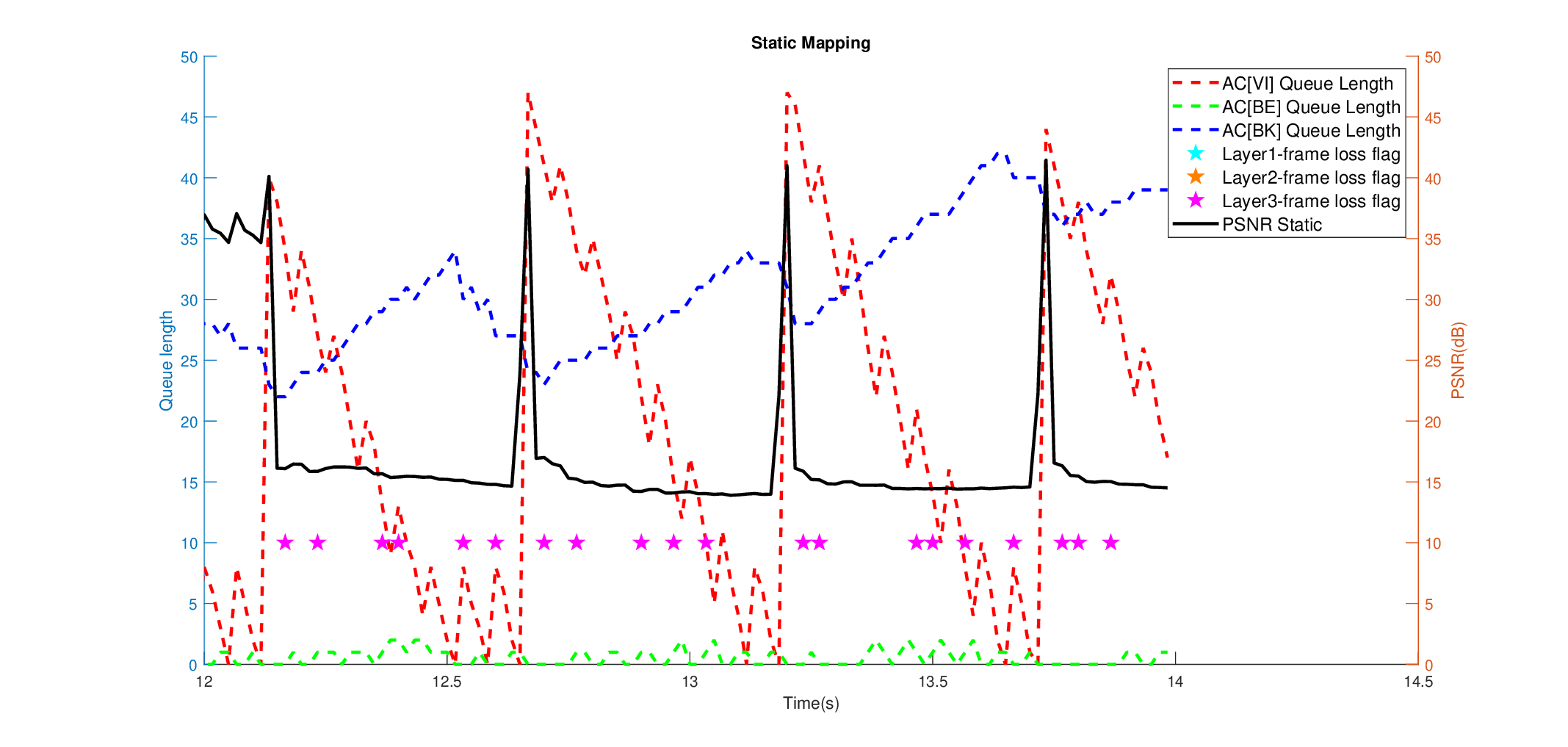}
\begin{figure*}[!htbp]
	\centering \makeatletter\IfFileExists{images/7.eps}{\includegraphics[width=1.03\linewidth]{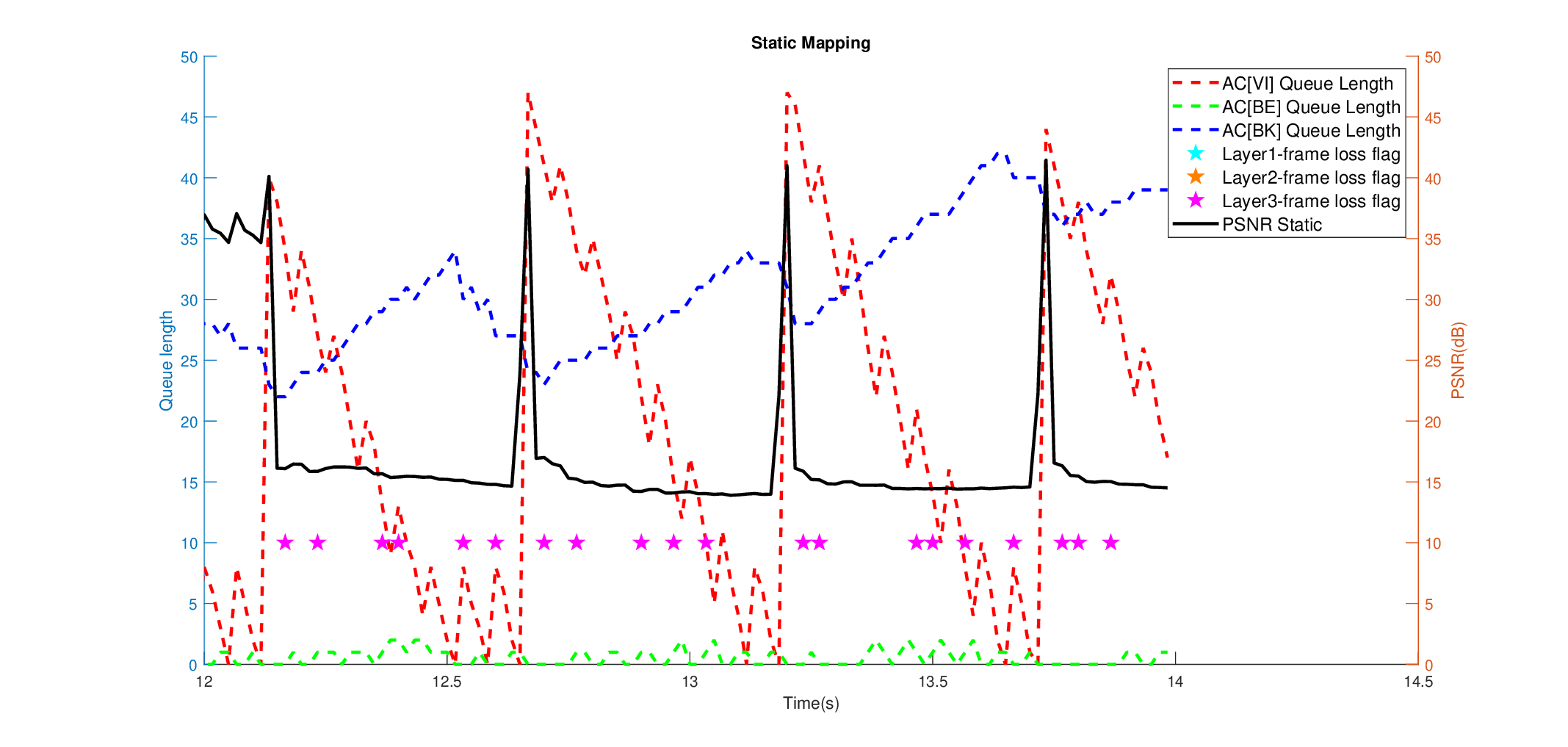}}{}
	\makeatother 
	\caption{{Static mapping algorithm: filling state of the queue, PSNR video quality and indication of lost images over time. }}
	\label{figure-ae03c2e7f761d33c007bf6285f67d8d7}
\end{figure*}
\egroup

\bgroup
\fixFloatSize{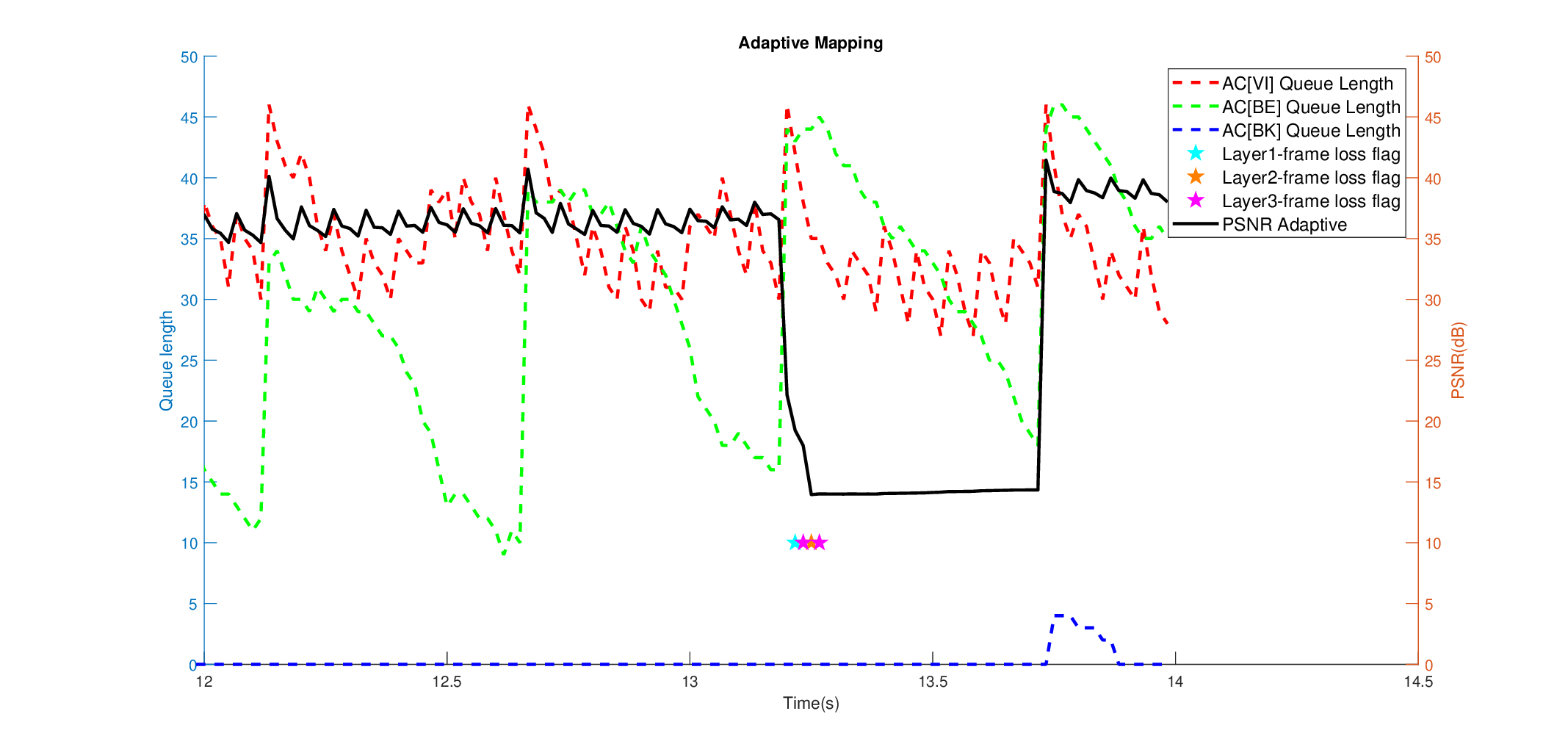}
\begin{figure*}[!htbp]
	\centering \makeatletter\IfFileExists{images/13.eps}{\includegraphics[width=1.03\linewidth]{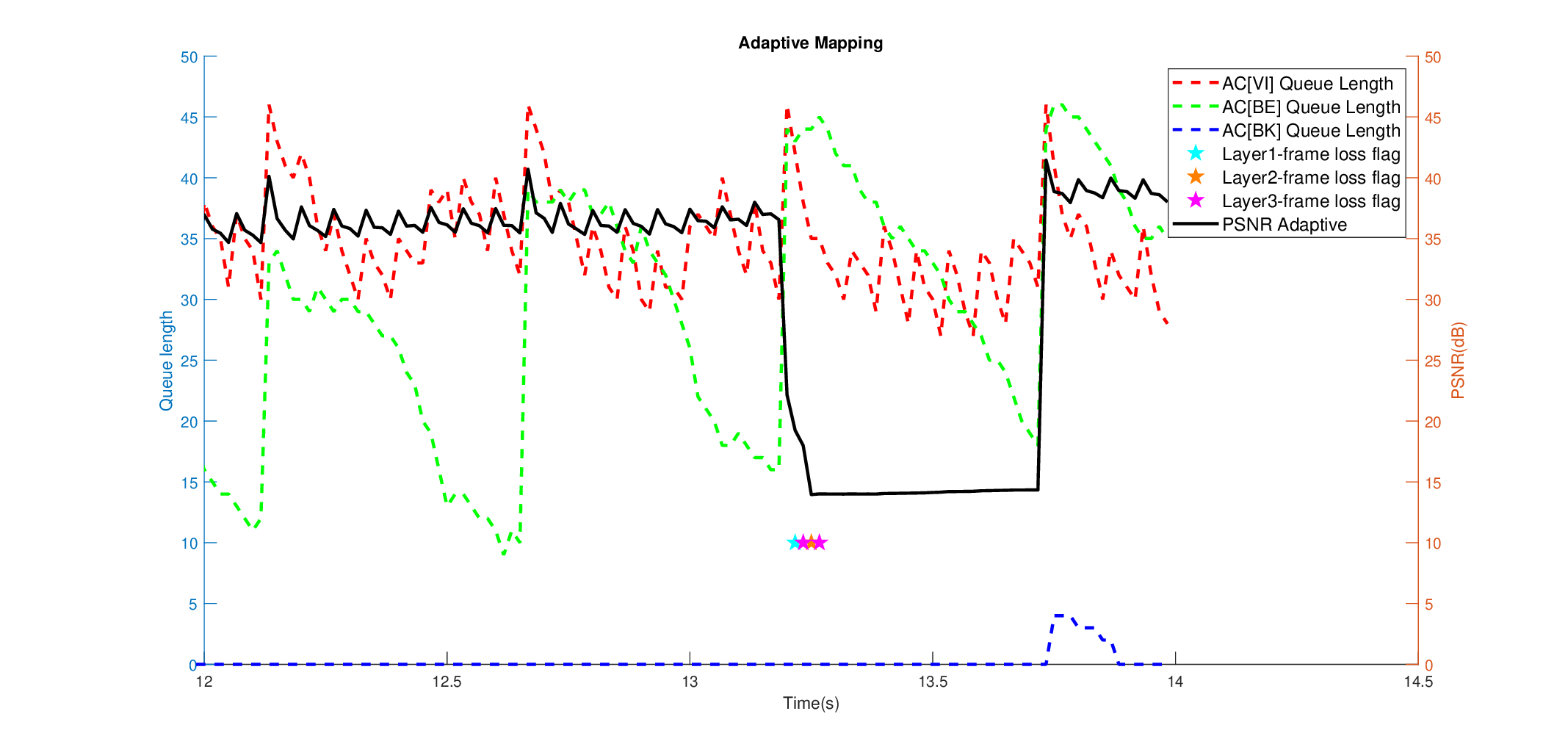}}{}
	\makeatother 
	\caption{{Adaptive mapping algorithm: filling state of the queue, PSNR video quality and indication of lost images over time. }}
	\label{figure-0ba3af5bc2f4a2b42adf4b64a82d9b2a}
\end{figure*}
\egroup
The PSNR peaks values in Figure~\ref{figure-ae03c2e7f761d33c007bf6285f67d8d7} are due to the good reception of IDR (layer-1) frame packets. Nevertheless, transmitting video packets through the background queue generates image losses due to the delay in receiving the packet, especially in low latency applications. We also find that the highest priority queues are not fully exploited as opposed to the adaptive method shown in Figure~\ref{figure-0ba3af5bc2f4a2b42adf4b64a82d9b2a}. Indeed, the queues filling in the adaptive method is done in a better suited way that reduces the lost packets number. This also explains the good PSNR values. A channel quality deterioration has caused a significant delay at the reception. This leads to the loss of four consecutive images of different layers. However, the loss in a layer-1 frame affects the PSNR during the entire GoP.

\subsection{ The efficiency of different mapping algorithms in a multi-stream transmission}In reality, the video stream usually coexists with other types of streams. This causes a network load and also causes VO, BE and BK ACs filled. In this case, the video packet access to the medium is more competitive. Also, for our methods, when video packets are mapped to lower priority queues, video packets share these resources with the dedicated packets. 

In order to study the mapping algorithms efficiency, we simulated the video transmission in our second scenario (highway environment + multi-stream). The results are established in Table~\ref{table-wrap-46e3b182c853081c3175e10e1fbaca8d}, several inferences can be made on different aspects of the transmission chain:

\begin{itemize}
  \item \relax Video sequence effect: Despite the same encoding rate between the different sequences, the encoding quality of the different sequences is obviously different. This is due to both the characteristics as well as the frame rate of each sequence. For example, the difference between Partyscene and BasketballDrill sequences encoding causes an average PSNR difference of 5.62 dB for the 3 configurations. Since the frame rate is the same (50 fps), the difference is due to the spatiotemporal characteristics specific to each sequence.
  \item \relax Video encoding configuration: We verify that encoding at the same bit rate with AI configuration logically results in a lower video quality than the other two coding modes. The RA and LD configurations have a fairly similar video quality with a slight advantage for the RA which is due to the forward and backward reference pictures of RA prediction structure.
  \item \relax Mapping Algorithm: For all video sequences, we verify an improvement in received packets number according to the efficiency of the mapping method. A comparison allows us to evaluate an average gain of 4\% between the EDCA algorithm and the static method. While the adaptive solution brings an average gain of 13\% compared to the static solution. This is also reflected in the video quality at the reception, the static solution brings an average PSNR gain of 1.21 dB compared to the EDCA algorithm. While the adaptive solution has an average PSNR gain of 8.74 dB compared to the static solution.
\end{itemize}
  
\begin{table*}[!htbp]
\caption{{Average PSNR and number of packet lost for each simulation.} }
\label{table-wrap-46e3b182c853081c3175e10e1fbaca8d}
\def\arraystretch{1}
\ignorespaces 
\centering 
\resizebox{\textwidth}{!}{%
\begin{tabulary}{\linewidth}{cclclll}
	\hline
	\multicolumn{1}{l}{\begin{tabular}[c]{@{}l@{}}Video \\ sequence\end{tabular}}                                        & \multicolumn{1}{l}{\begin{tabular}[c]{@{}l@{}}Coding \\ structure\end{tabular}} & \begin{tabular}[c]{@{}l@{}}Mapping \\ Algorithm\end{tabular} & \multicolumn{1}{l}{\begin{tabular}[c]{@{}l@{}}Encoding \\ PSNR\end{tabular}} & \begin{tabular}[c]{@{}l@{}}Transmission \\ PSNR\end{tabular} & \begin{tabular}[c]{@{}l@{}}Transmit \\ packet\end{tabular} & \begin{tabular}[c]{@{}l@{}}Received \\ packet\end{tabular} \\
	\tblmidrule
	\multirow{9}{*}{\begin{tabular}[c]{@{}c@{}}RaceHorses (30fps) \\ 832 x 480 (class C)\\ 300 frames\end{tabular}}      & \multirow{3}{*}{AI}                                                             & EDCA                                                         & \multirow{3}{*}{30.18}                                                       & 21.6                                                         & \multirow{3}{*}{2097}                                      & 1700                                                       \\
	&                                                                                 & Static mapping                                               &                                                                              & 21.6                                                         &                                                            & 1700                                                       \\
	&                                                                                 & Adaptive mapping                                             &                                                                              & 30.18                                                        &                                                            & 2097                                                       \\
	&
	\multirow{3}{*}{LD}                                                             & EDCA                                                         & \multirow{3}{*}{34.23}                                                       & 18.99                                                        & \multirow{3}{*}{2010}                                      & 1707                                                       \\
	&                                                                                 & Static mapping                                               &                                                                              & 20.73                                                        &                                                            & 1763                                                       \\
	&                                                                                 & Adaptive mapping                                             &                                                                              & 32.38                                                        &                                                            & 1994                                                    \\
	& \multirow{3}{*}{RA}                                                             & EDCA                                                         & \multirow{3}{*}{\textbf{34.4}}                                                        & 20.96                                                        & \multirow{3}{*}{2011}                                      & 1669                                                       \\
	&                                                                                 & Static mapping                                               &                                                                              & 21.4                                                         &                                                            & 1677                                                       \\
	&                                                                                 & Adaptive mapping                                             &                                                                              & \textbf{33.8}                                                         &                                                            & 1949                                                       \\
	\tblmidrule
	\multirow{9}{*}{\begin{tabular}[c]{@{}c@{}}PartyScene (50fps) \\ 832 x 480 (class C)\\ 500 frames\end{tabular}}      & \multirow{3}{*}{AI}                                                             & EDCA                                                         & \multirow{3}{*}{23.32}                                                       & 22.72                                                        & \multirow{3}{*}{2000}                                      & 1664                                                       \\
	&                                                                                 & Static mapping                                               &                                                                              & 22.72                                                        &                                                            & 1664                                                       \\
	&                                                                                 & Adaptive mapping                                             &                                                                              & 23.32                                                        &                                                            & 2000                                                       \\
	& \multirow{3}{*}{LD}                                                             & EDCA                                                         & \multirow{3}{*}{30.58}                                                       & 19.41                                                        & \multirow{3}{*}{2226}                                      & 1754                                                       \\
	&                                                                                 & Static mapping                                               &                                                                              & 20.73                                                        &                                                            & 1874                                                       \\
	&                                                                                 & Adaptive mapping                                             &                                                                              & \textbf{30.25}                                                        &                                                            & 2126                                                       \\
	& \multirow{3}{*}{RA}                                                             & EDCA                                                         & \multirow{3}{*}{\textbf{31.27}}                                                       & 21.15                                                        & \multirow{3}{*}{2233}                                      & 1712                                                       \\
	&                                                                                 & Static mapping                                               &                                                                              & 20.84                                                        &                                                            & 1778                                                       \\
	&                                                                                 & Adaptive mapping                                             &                                                                              & 29.63                                                        &                                                            & 2053                                                       \\
	\tblmidrule
	\multirow{9}{*}{\begin{tabular}[c]{@{}c@{}}BasketballDrill (50fps) \\ 832 x 480 (class C)\\ 500 frames\end{tabular}} & \multirow{3}{*}{AI}                                                             & EDCA                                                         & \multirow{3}{*}{29.1}                                                        & 26.4                                                         & \multicolumn{1}{c}{\multirow{3}{*}{2000}}                  & 1693                                                       \\
	&                                                                                 & Static mapping                                               &                                                                              & 26.4                                                         & \multicolumn{1}{c}{}                                       & 1693                                                       \\
	&                                                                                 & Adaptive mapping                                             &                                                                              & 29.1                                                         & \multicolumn{1}{c}{}                                       & 2000                                                       \\
	& \multirow{3}{*}{LD}                                                             & EDCA                                                         & \multirow{3}{*}{36.28}                                                       & 21.49                                                        & \multicolumn{1}{c}{\multirow{3}{*}{2082}}                  & 1709                                                       \\
	&                                                                                 & Static mapping                                               &                                                                              & 21.87                                                        & \multicolumn{1}{c}{}                                       & 1877                                                       \\
	&                                                                                 & Adaptive mapping                                             &                                                                              & \textbf{35.41}                                                       & \multicolumn{1}{c}{}                                       & 2081                                                       \\
	& \multirow{3}{*}{RA}                                                             & EDCA                                                         & \multirow{3}{*}{\textbf{36.67}}                                                       & 20.88                                                        & \multicolumn{1}{c}{\multirow{3}{*}{2069}}                  & 1700                                                       \\
	&                                                                                 & Static mapping                                               &                                                                              & 29.83                                                        & \multicolumn{1}{c}{}                                       & 1915                                                       \\
	&                                                                                 & Adaptive mapping                                             &                                                                              & 35.03                                                        & \multicolumn{1}{c}{}                                       & 2057                                                       \\
	\tblmidrule
	\multirow{9}{*}{\begin{tabular}[c]{@{}c@{}}BQMall (60fps) \\ 832 x 480 (class C)\\ 600 frames\end{tabular}}          & \multirow{3}{*}{AI}                                                             & EDCA                                                         & \multirow{3}{*}{27.7}                                                        & 21.04                                                        & \multicolumn{1}{c}{\multirow{3}{*}{2399}}                  & 1753                                                       \\
	&                                                                                 & Static mapping                                               &                                                                              & 21.04                                                        & \multicolumn{1}{c}{}                                       & 1753                                                       \\
	&                                                                                 & Adaptive mapping                                             &                                                                              & 26.9                                                         & \multicolumn{1}{c}{}                                       & 2392                                                       \\
	& \multirow{3}{*}{LD}                                                             & EDCA                                                         & \multirow{3}{*}{35.92}                                                       & 18.72                                                        & \multicolumn{1}{c}{\multirow{3}{*}{2169}}                  & 1768                                                       \\
	&                                                                                 & Static mapping                                               &                                                                              & 19.8                                                         & \multicolumn{1}{c}{}                                       & 1956                                                       \\
	&                                                                                 & Adaptive mapping                                             &                                                                              & \textbf{34.17}                                                        & \multicolumn{1}{c}{}                                       & 2169                                                       \\
	& \multirow{3}{*}{RA}                                                             & EDCA                                                         & \multirow{3}{*}{\textbf{36.47}}                                                       & 20.01                                                        & \multicolumn{1}{c}{\multirow{3}{*}{2203}}                  & 1770                                                       \\
	&                                                                                 & Static mapping                                               &                                                                              & 20.95                                                        & \multicolumn{1}{c}{}                                       & 1960                                                       \\
	&                                                                                 & Adaptive mapping                                             &                                                                              & 31.07                                                        & \multicolumn{1}{c}{}                                       & 2174         \\                                             	
\tblbottomrule 
\end{tabulary}}\par 
*AI: All Intra, LD: Low Delay, RA: Random Access.
\end{table*}

\subsection{Delay evaluation}Finally, given the low-latency transmissions importance in the context of vehicular networks, we were also interested in a delay evaluation of our different schemes. It is clear that the AI coding scheme offers the lowest calculation complexity (no estimation / motion compensation). It is also best suited to meet low latency constraints. Unfortunately, this is at the expense of coding distortion due to limited AI coding efficiency. The two other configurations are of better quality but with a higher latency. The following equations \unskip~\cite{259927:5821233} are defined for end-to-end delay calculation respectively for LD and RA encoding configuration:

\begin{equation}
{t_{low - delay}} = {t_{en - LD}} + {t_{net}} + {t_{dec - LD}}\ 
\label{moneq3}
\end{equation}
\begin{equation}
{t_{Random - Access - gop}} = \left( {gop - 1} \right) \times {t_{fr}} + \left( {{{\log }_2}\left( {gop} \right) + 1} \right) \times {t_{en - RA}} + {t_{net}} + {t_{dec - RA}}\ 
\label{moneq4}
\end{equation}

where: 

\begin{itemize}
	\item \relax \textit{t\ensuremath{_{\rm net}}} is the transmission/propagation delay, i.e. the delay time between the emitter and the receiver in a network system or architecture,
	\item \relax \textit{t\ensuremath{_{\rm en\ }}}the mean computational time for the encoding of one frame,
	\item \relax \textit{t\ensuremath{_{\rm dec}}} the mean computational time for the decoding of one frame. 
\end{itemize}
The GoP size is denoted as \textit{gop} and \textit{t\ensuremath{_{\rm fr}}} is the interval time between successive frames where \textit{t\ensuremath{_{\rm fr}}} is 1/\textit{fps}.

In order to estimate the HEVC encoding times for the AI, RA and LD configurations, a personal Intel Core i5-3210M 2.5 GHz PC with 6 GB of RAM was used. The computational resources were only dedicated to the encoding process. However, this produces significant encoding times. We assume that vehicles have the possibility of being able to have more powerful embedded encoders than the used PC, and that the parallel processing will reduce the computational cost. 

Moreover, to guarantee the real-time constraints, it is obvious that \textit{t\ensuremath{_{\rm en}} \textless\ t\ensuremath{_{\rm fr}}}\ensuremath{_{\rm }} must be satisfied for each frame \unskip~\cite{259927:5821234}. Simulation experiments showed that the LD configuration takes the most encoding time. RA and AI configurations are respectively 23\% and 72\% faster than the LD. Almost the same scores can be established if the encoding times of \unskip~\cite{259927:5821233} and \unskip~\cite{bossen_hevc_2012} are considered.

In order to evaluate the end-to-end delay in~\eqref{moneq3} and~\eqref{moneq4}, we vary the encoding time \textit{t\ensuremath{_{\rm en}}} from ${{t}_{0}}=2\times {{10}^{-3}}$ to \textit{t\ensuremath{_{\rm f}}}\ensuremath{_{\rm r}} (s). Since the encoding times of the different configurations are not equal, we also apply the previously described coefficient to the various encoding times.\mbox{}\protect\newline For example, if we take the BQmall sequence case. We vary t\ensuremath{_{\rm en\_LD\ }} from \textit{t\ensuremath{_{\rm 0}}} to (1/60) s and t\ensuremath{_{\rm en\_RA\ }} from $\alpha \times{{t}_{0}}$ to $\alpha \times(1/60)$ s. $\alpha$=1-0.23 comes from the experimental calculated encoding time difference. Moreover, the "Evalvid" tool-set provides an estimate value of \textit{t\ensuremath{_{\rm net}}}.

Thus, the end-to-end delay simulation evaluations of the different structures shows that the t\ensuremath{_{\rm low\_delay}} is 65\% faster than the t\ensuremath{_{\rm random\_access}}. The GoP size influence is undeniable, the transition to a GoP at 32 in the HM.16.16 used in this study (16 before) rules out the use of RA in a low delay application. A smaller GoP size could allow the use of RA in a low latency context as established by \unskip~\cite{259927:5821233}.
    
\section{Conclusion}
Video transmission in a vehicular environment is affected by various forms of losses, which results in packet loss and greatly affects perception of perceived quality. Moreover, it is obvious that the transmission of the video stream in real time over VANET is a quite challenging process. On the other hand, the new HEVC coder presents more promising results compared to its predecessor and allows significant advances in video coding in wireless environment. In this paper, an efficient adaptive cross-layer mapping algorithm has been presented to improve low latency HEVC streaming over IEEE 802.11p vehicular networks. The proposed improvement uses at the MAC layer application layer information in a cross-layer scheme. Indeed, information on temporal prediction video structure, frame type, and MAC layer buffer filling state, allow the algorithm to optimally delivery video packets.\mbox{}\protect\newline Several simulation results have shown a better overall performance improvement of the two proposed solutions compared to the standard EDCA, for different scenarios. Furthermore, a comparative study of QoS and QoE demonstrated that the proposed adaptive algorithm delivers the best results for the different HEVC temporal prediction structures.

The current solution does not bring any classification to the AI encoding configuration. Therefore, our next objective would be to consider a better video packets processing with such transmission. Also, with the delay constraint, packets that are not receipted within the time allocated are not taken into consideration. Thereby, their transmission is useless and only load the network. An algorithm capable of eliminating them at the transmitter will improve the transmission and deserves to be taking into consideration. The algorithm must establish a relation between the application delay constraint, the end-to-end delay and the buffering time represented in the queues filling.
   
%\section*{References}

\bibliographystyle{elsarticle-num}

\bibliography{article}

\end{document}